\PassOptionsToPackage{table}{xcolor}
\documentclass{hustmedia-paper}
\usepackage{amsmath,amssymb,booktabs,array,graphicx}
\usepackage{enumitem,flafter,placeins,float,needspace}

\definecolor{tblHeader}{HTML}{EAF0F7}
\definecolor{tblInk}{HTML}{263A50}
\definecolor{tblGroup}{HTML}{637081}
\definecolor{tblBaseline}{HTML}{F2F3F5}
\definecolor{tblOurs}{HTML}{EAE5F5}
\definecolor{tblSecond}{HTML}{FFF2C7}
\definecolor{tblOursInk}{HTML}{513B78}
\definecolor{tblGain}{HTML}{176B44}
\definecolor{tblLoss}{HTML}{A83E3E}
\definecolor{tblGainFill}{HTML}{E5F2E9}
\definecolor{tblLossFill}{HTML}{FAEAEA}
\newcommand{\tblhead}[1]{\textcolor{tblInk}{\textbf{#1}}}

\newlength{\tblValueWidth}
\newlength{\tblMeasureWidth}

\setcopyright{none}
\setcitestyle{authoryear,round,citesep={;},aysep={,},yysep={;}}
\newcommand{\method}{RSRL}
\newcommand{\E}{\mathbb{E}}
\newcommand{\clip}{\operatorname{clip}}

\newcommand{\sg}{\operatorname{stopgrad}}

\title[Rubrics Beyond Evaluation]{Rubrics Beyond Evaluation: Exploiting Rubric Structure in Reinforcement Learning}
\HMMDisplayTitlePair{Rubrics Beyond Evaluation:}{Exploiting Rubric Structure in Reinforcement Learning}
\author{Yuxun Huang}
\affiliation{
  \institution{Huazhong University of Science and Technology}
  \city{Wuhan}
  \country{China}
}
\email{yuxun_huang@hust.edu.cn}
\author{Yunyao Zhang}
\affiliation{
  \institution{Huazhong University of Science and Technology}
  \city{Wuhan}
  \country{China}
}
\email{ikostar@hust.edu.cn}
\author{ZeLiang Chen}
\affiliation{
  \institution{Huazhong University of Science and Technology}
  \city{Wuhan}
  \country{China}
}
\email{zeliang_chen@hust.edu.cn}
\author{Yi-Ping Phoebe Chen}
\affiliation{
  \institution{La Trobe University}
  \city{Melbourne}
  \country{Australia}
}
\email{phoebe.chen@latrobe.edu.au}
\author{Junqing Yu}
\affiliation{
  \institution{Huazhong University of Science and Technology}
  \city{Wuhan}
  \country{China}
}
\email{yjqing@hust.edu.cn}
\author{Zikai Song}
\correspondingauthor
\affiliation{
  \institution{Huazhong University of Science and Technology}
  \city{Wuhan}
  \country{China}
}
\email{skyesong@hust.edu.cn}
\HMMAuthorLine{\href{https://openreview.net/profile?id=~Yuxun_Huang1}{Yuxun Huang}\textsuperscript{1,*} \textbullet\ \href{https://openreview.net/profile?id=~Yunyao_Zhang1}{Yunyao Zhang}\textsuperscript{1,*} \textbullet\ \href{https://openreview.net/profile?id=~ZeLiang_Chen2}{ZeLiang Chen}\textsuperscript{1}\\[2pt]\href{https://openreview.net/profile?id=~Yi-Ping_Phoebe_Chen1}{Yi-Ping Phoebe Chen}\textsuperscript{2} \textbullet\ \href{https://openreview.net/profile?id=~Junqing_Yu1}{Junqing Yu}\textsuperscript{1} \textbullet\ \href{https://openreview.net/profile?id=~Zikai_Song1}{Zikai Song}\textsuperscript{1,\ensuremath{\dagger}}}
\HMMAffiliationLine{\textsuperscript{1}Huazhong University of Science and Technology, Wuhan, China\\[1pt]\textsuperscript{2}La Trobe University, Melbourne, Australia}
\HMMEqualContribution{Yuxun Huang and Yunyao Zhang contributed equally to this work.}
\HMMCorrespondingAuthor{Zikai Song at \HMMEmail{skyesong@hust.edu.cn}.}
\HMMCorrespondingEmail{skyesong@hust.edu.cn}
\HMMPublicationStatus{Preprint}
\HMMAbstractEmailOff
\HMMAuthorInformation{
  Yuxun Huang\textsuperscript{1} (\HMMEmail{yuxun_huang@hust.edu.cn});
  Yunyao Zhang\textsuperscript{1} (\HMMEmail{ikostar@hust.edu.cn});
  ZeLiang Chen\textsuperscript{1} (\HMMEmail{zeliang_chen@hust.edu.cn});
  Yi-Ping Phoebe Chen\textsuperscript{2} (\HMMEmail{phoebe.chen@latrobe.edu.au});
  Junqing Yu\textsuperscript{1} (\HMMEmail{yjqing@hust.edu.cn});
  Zikai Song\textsuperscript{1} (\HMMEmail{skyesong@hust.edu.cn}).
}
\HMMAuthorInformationOn
\newcommand{\PaperPDFAuthors}{Yuxun Huang, Yunyao Zhang, ZeLiang Chen, Yi-Ping Phoebe Chen, Junqing Yu, Zikai Song}

\AtBeginDocument{}

\hypersetup{pdfauthor={\PaperPDFAuthors}}
\begin{abstract}
Rubrics are widely used to evaluate open-ended question answering by breaking overall quality into explicit criteria \citep{kim2024prometheus,arora2025healthbench}. These explicit criteria make rubrics a rich source of learning guidance, yet their potential for guiding the training process has been largely overlooked. Existing methods use rubrics mainly as direct feedback signals, without explicitly modeling the capabilities behind their criteria \citep{gunjal2025rubrics,li2026rubrichub}. This leaves their capability structure largely unused in guiding the training process. We introduce Rubric-Structured Reinforcement Learning (RSRL), a RL framework that unlocks the full potential of rubrics by exploiting their structure throughout the training process, rather than treating them simply as direct reward signals. At its core, RSRL builds on three key designes: (1) \textbf{\textit{Rubric-to-Capability Structuring}} first recovers the capability structure hidden in rubric criteria, (2) \textbf{\textit{Self-Sufficient Capability Learning}} then uses this structure to bootstrap underdeveloped capabilities into effective learning, (3) \textbf{\textit{Capability-Residual Credit Routing}} finally preserves capability-specific gains during optimization. Experiments show that RSRL consistently improves both overall performance and underdeveloped capabilities, boosting downstream scientific question-answering performance by up to 7.43 over the initial policy and 7.60 over non-rubric structured RL baselines, showing that reinforcement learning can substantially benefit from the capability structure encoded in rubrics.

\end{abstract}
\begin{document}
\maketitle
\Needspace{6\baselineskip}
\section{Introduction}
\textbf{\textit{Rubrics}} \citep{kim2024prometheus} have become increasingly popular for evaluating complex model outputs across a wide range of open-ended tasks \citep{arora2025healthbench,yifei2025researchqa}, as they provide a structured way to break overall quality into explicit criteria. These criteria can capture aspects such as correctness, completeness, evidence use, reasoning quality, and clarity. More importantly, these criteria are not merely scoring rules, they reveal the capabilities required to produce high-quality responses, making rubrics \textbf{\textit{a rich source of guidance for model learning}}. However, this potential has barely been explored. Most existing work \citep{kim2024prometheus,arora2025healthbench} uses rubrics only for evaluation, and the few attempts (\citealp{li2026rubrichub}; \citealp{gunjal2025rubrics,shao2026drtulu}) that incorporate them into reinforcement learning, they largely treat individual criteria as direct supervision or reward signals. As a result, the capability structure encoded across rubric criteria \textbf{\textit{remains largely unused}} during learning.

We argue that rubrics should not be treated merely as collections of reward criteria, but as structured descriptions of the capabilities required to solve a task \citep{zhang-etal-2026-semantic,zheng2026surfacing}. The key is to preserve and exploit this structure throughout reinforcement learning, rather than collapsing it into a single aggregate training signal. This shifts rubric-based learning \textbf{\textit{from criterion-level optimization to capability-level development}}: rubric criteria reveal what capabilities the policy lacks, and the capability structure can further guide how learning resources are allocated during training \citep{li2026lora}. Based on this view, we formulate \textbf{Rubric-Structured Reinforcement Learning (RSRL)}, which turns rubrics from an evaluation tool into a structure that actively shapes the learning process \citep{song2026socialintelligence}.

Three key challenges arise when bringing rubric structure into reinforcement learning: \textbf{(C1) \textit{What capability is underdeveloped?}} Rubric criteria are defined at the question level, while the weaknesses they reveal may reflect shared capabilities across many questions, making it difficult to recover these shared capabilities and identify which ones remain weak; \textbf{(C2) \textit{How can weak capabilities be effectively explored?}} Once a weak capability is identified, natural rollouts may still provide too little useful variation for the policy to discover better behaviors, making it difficult to achieve effective early-stage improvement. \textbf{(C3) \textit{How can capability gains be preserved?}} Even when better capability-specific behaviors are found, their signals can be diluted once multiple rubric criteria are aggregated during optimization, causing the improvement to have little effect on the final policy update.

To address these challenges, our RSRL carries rubric structure through the full learning process rather than using it only at the reward level. (For \textbf{C1}) we introduce \textbf{Rubric-to-Capability Structuring} to recover the capability structure hidden in rubric criteria. Specifically, we construct Hierarchical Capability Channels to organize fragmented criteria into shared capabilities and track which capabilities remain underdeveloped across questions. \textbf{(For C2), Self-Sufficient Capability Learning} is designed to bootstrap underdeveloped capabilities when the current policy lacks useful learning behaviors, while allowing them to transition toward independent learning as they improve \citep{song2023compact}. To realize this idea, we develop a Self-Sufficiency Gating Module that introduces capability-enhanced rollouts through capability-specific prompting when additional support is needed, and adaptively withdraws them once the capability becomes sufficiently established. (For \textbf{C3}), we further introduce \textbf{Capability-Residual Credit Routing} to preserve capability-specific gains during aggregate optimization. In practice, we formulate a Capability Residual Projection Policy to separate the target capability's residual improvement from the aggregate advantage and carry this signal into the final policy update. Experiments on ResearchQA \citep{yifei2025researchqa}, closed-book SQAv2 \citep{bragg2026astabench}, and GPQA-Diamond \citep{rein2024gpqa} show that RSRL improves downstream scientific question-answering performance by up to 11.52\% over the initial policy and 11.81\% over non-rubric RL baselines. It also achieves the best final worst-capability and overall response scores on a fixed 128-question development probe, while preserving gains in strong capabilities. These results demonstrate the value of exploiting rubric-encoded capability structure throughout reinforcement learning.

Our main contributions are as follows:
\begin{itemize}[leftmargin=*,itemsep=2pt,topsep=3pt]
\item \textbf{A systematic formulation of rubric-structured reinforcement learning.} We are the first to deeply examine how rubric structure can be exploited beyond evaluation and direct reward supervision, revealing its role in capability modeling, learning allocation, and policy optimization. This establishes RSRL as a general framework for integrating rubric structure throughout reinforcement learning.
\item \textbf{A structured RL framework built around capability development.} We instantiate this view with RSRL through three coordinated designs: \textit{Rubric-to-Capability Structuring} recovers shared capability structure from rubric criteria, \textit{Self-Sufficient Capability Learning} enables efficient improvement of underdeveloped capabilities, and \textit{Capability-Residual Credit Routing} preserves capability-specific gains during policy optimization.
\end{itemize}

\Needspace{6\baselineskip}
\section{Related Work}
\paragraph{Rubric-based supervision and exploration.}
Rubrics as Rewards uses instance-specific criteria as RL feedback \citep{gunjal2025rubrics}; RubricHub scales rubric construction and rubric-based post-training \citep{li2026rubrichub}. DR Tulu evolves criteria with the policy using newly retrieved evidence and contrasting responses \citep{shao2026drtulu}. RuscaRL guides exploration with rubric scaffolds that decay over training \citep{zhou2026ruscarl}. RISE-RL targets repeatedly missed criteria and learns from reward-filtered guided trajectories through an auxiliary objective under the original prompt \citep{hou2026rise}. CRAFT builds a hierarchical capability tree from rubric criteria to guide targeted supervised fine-tuning \citep{gupta2026craft}. The RL methods above organize reinforcement around question-specific criteria, without a shared capability diagnosis that jointly coordinates online exploration and response-level credit across questions.

\paragraph{Adaptive allocation and reward balancing.}
Automated curricula allocate training exposure using learning progress \citep{graves2017curriculum,zhang-etal-2026-logical}. For multi-reward optimization, GDPO normalizes reward dimensions separately \citep{liu2026gdpo}, GD$^2$PO filters conflicting rollouts and reweights queries by reward agreement \citep{liu2026gd2po}, and SMOPD distills reward-specialized teachers into one policy \citep{wang2026smopd}. These methods manage training exposure and reward trade-offs, but do not specify how persistent capability deficits should jointly determine exploration support and response-level credit.

\paragraph{Controlled generation and sampling correction.}
PREADD contrasts distributions with and without a control prefix \citep{pei2023preadd}, while contrastive decoding compares stronger and weaker models \citep{li2023contrastive}. Multiple importance sampling provides density-based weights for combining samples from different proposals through the balance heuristic \citep{veach1995optimally}. These tools support generation control and sampling correction, but leave unspecified when a weak capability needs guidance and when that guidance should be withdrawn during learning.

Together, these gaps motivate a learning process organized around evolving capability needs. RSRL uses rubric structure to connect shared capability diagnosis, selective exploration, and capability-specific policy updates, making the policy's diagnosed capability state govern the provision and withdrawal of exploration support.

\Needspace{6\baselineskip}
\section{Problem Motivation}
\label{sec:problem_motivation}
Rubric-based RL needs sampled responses that provide useful variation on weak capabilities. When unguided rollouts rarely satisfy a requirement, rubric feedback can identify the gap without supplying successful behaviors to learn from. Reweighting scores alone cannot introduce behaviors absent from the sampled group. Self-Sufficient Capability Learning uses capability-guided proposals when diagnostics identify insufficient support and a sufficiently large deficit, then withdraws guidance when its activation conditions no longer hold \citep{zhang2026intervensiminterventionawaresocialnetwork}. Guided responses are learned under the original question context, with sampling weights accounting for the proposal shift.

RSRL connects this exploration process to diagnosis and optimization through a shared capability target. Rubric-to-Capability Structuring groups question-local criteria into cross-question channels \citep{song2026hypergraph} and combines current evidence with history \citep{song2025temporal,wang2026seeingwiderjointspatiotemporal,zhang2026coupling}. Capability-Residual Credit Routing carries target-specific distinctions into response-level advantages. Together, these stages coordinate which capability receives exploration support and how its feedback affects policy updates.
\par

\Needspace{6\baselineskip}
\section{Rubric-Structured Reinforcement Learning}
\label{sec:method}
\begin{figure}[tbp]
\centering
\includegraphics[width=\linewidth]{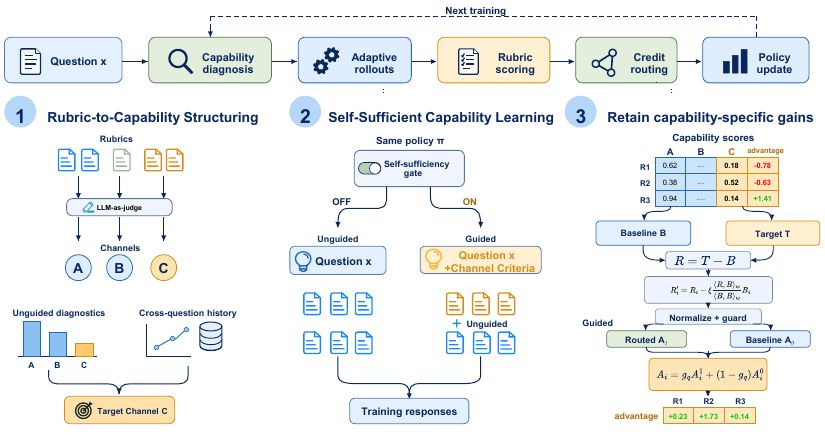}
\caption{Overview of \method{}. Rubric-to-Capability Structuring identifies targets from local and cross-question evidence. Self-Sufficient Capability Learning gates guided exploration. Capability-Residual Credit Routing adjusts response credit, with a guard selecting the routed candidate or baseline. Sampling weights account for the proposal shift; updates use the original question.}
\Description{RSRL links rubric-to-capability structuring, self-sufficient capability learning, and capability-residual credit routing.}
\label{fig:method}
\end{figure}

RSRL uses rubric structure to identify weak capabilities, guide exploration, and route capability-specific signals into policy updates under the original question (Figure~\ref{fig:method}).

\Needspace{5\baselineskip}
\subsection{Rubric-to-Capability Structuring}
\label{sec:capability_structuring}
Rubric-to-Capability Structuring uses \textbf{Hierarchical Capability Channels} to link question-specific rubric requirements with recurring domain capabilities.

\paragraph{Constructing capability channels.}
For a question $q$, let $x_q$ denote its original context, $\mathcal R_q$ its rubric criteria, and $\mathcal C_q$ its applicable local capability channels. Each source criterion is counted once and assigned to one local channel. A domain codebook $\mathcal Z$ links channels across questions through
\begin{equation}
m_q:\mathcal C_q\rightarrow\mathcal Z\cup\{\bot\},
\end{equation}
where $\bot$ denotes no reliable match. An LLM proposes the definitions, and an LLM judge assesses their correspondence. Definitions and mappings are frozen before training. Unmatched channels retain local diagnosis and training credit but do not update domain state (Appendix~\ref{app:channels}).

\paragraph{Estimating capability performance.}
Before update $s$, we sample unguided diagnostics from $\pi_{\theta_{s-1}}(\cdot\mid x_q)$ separately from optimization rollouts. Criterion scores lie in $[0,1]$, with higher values better. We average valid observations, including zeros, per criterion and combine them with fixed within-channel weights to obtain $S_{q,c}\in[0,1]$.

For a mapped channel, we pool current observations with cross-question history whenever both estimates are available:
\begin{equation}
\widetilde S_{q,c} = \nu_{q,c}S_{q,c} + (1-\nu_{q,c})U_{m_q(c),s-1},
\qquad \nu_{q,c} = \frac{n_{q,c}}{n_{q,c}+\kappa}.
\label{eq:diagnose}
\end{equation}
Here $U_{z,s-1}$ is the historical domain estimate, $n_{q,c}$ counts distinct responses with valid channel evidence, and $\kappa>0$ controls pooling. We use either estimate directly if the other is unavailable; channels lacking both are excluded from target selection.

\paragraph{Selecting the target capability.}
Channels with available estimates form $\mathcal C_q^{\mathrm{diag}}\subseteq\mathcal C_q$. A fixed performance threshold $\tau\in(0,1]$ defines their normalized deficit:
\begin{equation}
h_{q,c} = \max\left(0,\frac{\tau-\widetilde S_{q,c}}{\tau}\right).
\end{equation}
If $\mathcal C_q^{\mathrm{diag}}$ is empty, targeted exploration and residual routing are disabled for that question. Otherwise, we select the channel with the largest estimated deficit:
\begin{equation}
c_q^* = \arg\max_{c\in\mathcal C_q^{\mathrm{diag}}}h_{q,c},
\qquad h_q^*=h_{q,c_q^*}.
\label{eq:target}
\end{equation}
Stable channel identifiers break ties. If $h_q^*=0$, both exploration and routing coefficients are zero. Reliability $\chi_q\in[0,1]$ combines local evidence, domain-prior confidence, and mapping confidence for routing (Appendix~\ref{app:record}).

\paragraph{Maintaining the domain state.}
We initialize domain estimates with unguided diagnostics of the starting policy. At each update, observations sharing a domain capability are merged within questions and averaged equally across questions to obtain $\bar U_{z,s}$. After the optimizer step,
\begin{equation}
U_{z,s} = \beta_D U_{z,s-1} + (1-\beta_D)\bar U_{z,s},
\label{eq:domain_update}
\end{equation}
where $\beta_D\in[0,1)$ controls historical retention. The selected channel and deficit guide exploration and credit allocation in Sections~\ref{sec:mis}--\ref{sec:rar}.

\Needspace{5\baselineskip}
\subsection{Self-Sufficient Capability Learning}
\label{sec:mis}
Self-Sufficient Capability Learning uses a \textbf{Self-Sufficiency Gating Module} to activate capability-guided sampling while retaining the original question for policy updates.

\noindent\textbf{Deciding when guidance is needed.}
Guidance requires sufficient diagnostic evidence, a deficit above the activation threshold, and either uniformly minimal scores (\textbf{Zero}) or varying scores with few positive responses (\textbf{Low}). Other cases use unguided sampling. Reassessment before each update withdraws guidance whenever an activation condition fails, tying exploration support to the policy's current diagnostic state rather than a fixed training schedule. These are observed score patterns, not probability-support claims (Appendix~\ref{app:ctse}).

\noindent\textbf{Constructing guided proposals.}
At a shared response prefix $y_{<t}$, we evaluate three distributions from the same pre-update policy: $p_{0,t}$ conditions on $x_q$; $p_{c,t}$ adds the target capability description and training criteria; and $p_{n,t}$ adds nondirectional guidance. This target--control comparison provides a reference for distinguishing capability cues from the effects of extra prompting, adapting prefix-based control \citep{pei2023preadd}.

Using the full softmax distributions, we form the target--control contrast
\begin{equation}
u_t(v) = \log p_{c,t}(v)-\log p_{n,t}(v),
\end{equation}
where $v$ denotes a candidate token. We standardize this contrast under the unguided distribution and clip its magnitude:
\begin{equation}
\delta_t(v) = \operatorname{clip}\!\left(
\frac{u_t(v)-\mathbb E_{v'\sim p_{0,t}}\!\left[u_t(v')\right]}
{\sqrt{\operatorname{Var}_{v'\sim p_{0,t}}\!\left[u_t(v')\right]+\varepsilon}},
-b,b
\right).
\end{equation}
Here, $b>0$ bounds the standardized contrast, and $\varepsilon>0$ stabilizes the variance normalization. We apply the resulting perturbation to the unguided distribution:
\begin{equation}
q_{\eta,t}(v) =
\frac{p_{0,t}(v)\exp\!\left(\eta_q\delta_t(v)\right)}
{\sum_{v'}p_{0,t}(v')\exp\!\left(\eta_q\delta_t(v')\right)}.
\label{eq:proposal}
\end{equation}
The strength $\eta_q\geq0$ increases with $h_q^*$ up to a cap, with a smaller multiplier for \textbf{Low}. At $\eta_q=0$, the proposal is $p_{0,t}$. Appendix~\ref{app:ctse} specifies the schedule and prompt content.

\noindent\textbf{Learning under the original question.}
An active gate allocates optimization responses equally between guided and unguided sampling; otherwise, all are unguided.

Guided sampling can change the response distribution. We account for this shift using sequence-level importance weights based on the balance heuristic of \citet{veach1995optimally}. Let $P_0(y\mid x_q)$ denote the pre-update unguided sequence distribution. The guided sequence distribution is
\begin{equation}
Q_\eta(y\mid x_q) = \prod_{t=1}^{|y|}q_{\eta,t}(y_t),
\end{equation}
where each token probability is evaluated at the corresponding response prefix. The mixture density associated with the equal allocation is
\begin{equation}
M(y\mid x_q) = \tfrac12P_0(y\mid x_q) + \tfrac12Q_\eta(y\mid x_q).
\end{equation}
We evaluate both sequence densities for every optimization response, regardless of which proposal generated it. The raw importance ratios and self-normalized weights are
\begin{equation}
\omega_i = \frac{P_0(y_i\mid x_q)}{M(y_i\mid x_q)},
\qquad w_i = \frac{\omega_i}{\sum_{j=1}^{G}\omega_j},
\label{eq:mis}
\end{equation}
Here $G$ is the group size, and densities include each proposal's sampling and stopping rules. Since $M\geq P_0/2$, $0\leq\omega_i\leq2$. Raw ratios weight the loss; $w_i$ weights routing statistics. Unguided-only groups use $\omega_i=1$, $w_i=1/G$. Appendix~\ref{app:update} specifies density evaluation and overlap checks; Appendix~\ref{app:properties} gives the scope of correction.

\Needspace{5\baselineskip}
\subsection{Capability-Residual Credit Routing}
\label{sec:rar}
\label{sec:guard}
Capability-Residual Credit Routing uses a \textbf{Capability Residual Projection Policy} to retain target-specific distinctions relative to a channel-balanced baseline. It partially attenuates their shared component, scales the residual by diagnosed need, and applies a group-level guard.

\noindent\textbf{Constructing target-specific credit.}
We standardize valid criterion scores across the $G$ responses with unweighted moments to obtain $a_{ir}$; invalid judgments and inactive criteria contribute zero. Fixed within-channel weights $p_q(r\mid c)$ sum to one without redistribution when criteria become inactive. Channel advantages and the baseline are
\begin{equation}
C_{ic} = \sum_{r\mapsto c}p_q(r\mid c)a_{ir},
\qquad B_i = \frac{1}{|\mathcal C_q|}\sum_{c\in\mathcal C_q}C_{ic}.
\label{eq:credit}
\end{equation}
Appendix~\ref{app:credit} specifies criterion eligibility and standardization. For a diagnosed target, we compare its advantage with this baseline and form their residual difference:
\begin{equation}
T_i=C_{i,c_q^*}, \qquad R_i=T_i-B_i.
\end{equation}
\noindent\textbf{Projecting and scaling the residual.}
To control overlap with baseline credit, we use the self-normalized weights from Section~\ref{sec:mis} and define $\langle a,b\rangle_w=\sum_iw_i a_ib_i$ and $\|a\|_w=\sqrt{\langle a,a\rangle_w}$. We partially attenuate the baseline-aligned component and scale the resulting residual:
\begin{align}
R'_i &= R_i-\xi\frac{\langle R,B\rangle_w}{\langle B,B\rangle_w}B_i,
\label{eq:residual}\\
H_i &= B_i+\lambda_q r_{\mathrm{ref}}\frac{R'_i}{\|R'\|_w}.
\label{eq:rar}
\end{align}
Here, $0<\xi<1$ controls the degree of partial projection. We set $R'=R$ when baseline energy is negligible. The coefficient $\lambda_q=\chi_q f(h_q^*)$ links correction strength to the diagnosed deficit and reliability. The schedule $f$ is fixed, bounded, and nondecreasing, with $f(0)=0$. The positive reference scale $r_{\mathrm{ref}}$ is computed from the weighted norms of the baseline and channel advantages. Appendix~\ref{app:credit} gives both definitions.

When applied, the correction has weighted norm $\lambda_q r_{\mathrm{ref}}$. We set $H=B$ when routing is disabled, the question has only one channel, or the weighted residual norm is below a numerical tolerance.

\noindent\textbf{Selecting the final credit assignment.}
The guard checks the final normalized advantages used for policy optimization. Let $\mathcal N_w$ denote weighted advantage normalization with centering and magnitude control, as specified in Appendix~\ref{app:credit}. We compute
\begin{equation}
A^0=\mathcal N_w(B), \qquad A^1=\mathcal N_w(H), \qquad \Delta A=A^1-A^0.
\end{equation}
We distinguish target opportunities from response preferences shared with the baseline:
\begin{equation}
\begin{aligned}
\mathcal U &= \{(i,j):T_i>T_j,\ B_i\leq B_j\},\\
\mathcal P &= \{(i,j):B_i\geq B_j,\ T_i\geq T_j\}.
\end{aligned}
\end{equation}
We accept the candidate only when all four conditions hold:
\begin{equation}
\begin{aligned}
&\mathcal U\neq\varnothing, \qquad \langle T,\Delta A\rangle_w>0,\\
&A^1_i\geq A^1_j \quad\text{for every }(i,j)\in\mathcal P,\\
&\|\Delta A\|_w \leq \frac{\|R'\|_w}{r_{\mathrm{ref}}}.
\end{aligned}
\label{eq:guard}
\end{equation}
The guard requires a target opportunity and a target-aligned change, preserves shared orderings, and bounds the adjustment. With acceptance $g_q\in\{0,1\}$, select
\begin{equation}
A_i=g_qA^1_i+(1-g_q)A^0_i.
\label{eq:selected}
\end{equation}
No normalization follows selection; questions without a target use $A^0$. Appendix~\ref{app:guard} proves these within-group credit properties.

\noindent\textbf{Updating the policy.}
Each selected advantage weights the tokens of its response. We accumulate gradients over one batch pass and take one optimizer step. With proposal weights, judgments, and advantages held fixed, the implemented loss has the same local gradient as the following surrogate at $\theta=\theta_{s-1}$:
\begin{equation}
\mathcal L_{\mathrm{sp}}(\theta)
= \frac{1}{|\mathcal B|G}\sum_{q\in\mathcal B}\sum_{i=1}^{G}\omega_{qi}
\left[
-\frac{A_{qi}}{L_{qi}}\sum_{t=1}^{L_{qi}}
\log\pi_\theta\!\left(y_{qit}\mid x_q,y_{qi,<t}\right)
+\beta_{\mathrm{KL}}\bar k_{qi}(\theta)
\right].
\label{eq:loss}
\end{equation}
Here $\mathcal B$ is the question batch, $L_{qi}$ the response length, and $\omega_{qi}$ the raw proposal ratio. The token-averaged penalty $\bar k_{qi}$ compares the current and frozen starting policies with strength $\beta_{\mathrm{KL}}$. Appendix~\ref{app:update} gives the implemented loss.
\par

\Needspace{6\baselineskip}
\section{Experiments}
\label{sec:experiments}
We evaluate scientific question answering and capability development. Table~\ref{tab:main} summarizes benchmark performance; we then examine component contributions, capability changes, and guidance strength.

\begin{table}[tbp]
\centering
\caption{\textbf{Scientific question-answering scores.} Upper blocks provide external references; Qwen3-4B methods use our evaluation. Higher is better. Initial is the starting checkpoint. Parentheses are differences from Initial: score points, or percentage points for GPQA-Diamond. Within Qwen3-4B, bold/underlined scores mark best/second-best, including ties. Green/red parentheses mark gains/drops. Darker purple behind Ours indicates larger gains from Initial.}
\label{tab:main}

\renewcommand{\arraystretch}{1.15}
\setlength{\tabcolsep}{1.5pt}
\setlength{\aboverulesep}{0pt}
\setlength{\belowrulesep}{0pt}
\setlength{\extrarowheight}{0.5pt}
\newlength{\mainScoreWidth}
\newlength{\mainDeltaWidth}
\settowidth{\mainScoreWidth}{\textbf{88.88}}
\settowidth{\mainDeltaWidth}{(+0.00)}
\newcommand{\alignedscore}[2]{\makebox[\mainScoreWidth][r]{#1}\makebox[\mainDeltaWidth][l]{#2}}
\newcommand{\mainmetrichead}[1]{\makebox[\mainScoreWidth][c]{\textbf{#1}}\makebox[\mainDeltaWidth]{}}
\definecolor{mainGainPurple}{HTML}{8066CC}
\newcommand{\gainshadedscore}[3]{\begingroup\setlength{\fboxsep}{0pt}\colorbox{mainGainPurple!#1!white}{\strut\alignedscore{#2}{#3}}\endgroup}
\newcommand{\thickscoreunderline}[1]{\begingroup\setbox0=\hbox{#1}\leavevmode\rlap{\raisebox{-2pt}[0pt][0pt]{\rule{\wd0}{0.7pt}}}\box0\endgroup}
\begin{tabular}{>{\raggedright\arraybackslash}p{\dimexpr0.276\linewidth-10\tabcolsep\relax}*{4}{>{\centering\arraybackslash}p{0.181\linewidth}}}
\toprule[1pt]
\textbf{Method} & \mainmetrichead{ResearchQA} & \mainmetrichead{SQAv2} & \mainmetrichead{GPQA-Diamond} & \mainmetrichead{RubricHub} \\
\midrule[0.4pt]
\multicolumn{5}{l}{\textcolor{tblGroup}{Closed-source Frontier}} \\
GPT-5 & \alignedscore{77.60}{} & \alignedscore{---}{} & \alignedscore{85.70}{} & \alignedscore{---}{} \\
Claude Opus 4.1 & \alignedscore{74.30}{} & \alignedscore{86.10}{} & \alignedscore{77.30}{} & \alignedscore{---}{} \\
GPT-4.1 & \alignedscore{70.80}{} & \alignedscore{---}{} & \alignedscore{50.50}{} & \alignedscore{---}{} \\
DeepSeek V3.1 & \alignedscore{75.90}{} & \alignedscore{---}{} & \alignedscore{68.30}{} & \alignedscore{---}{} \\
\midrule[0.4pt]
\multicolumn{5}{l}{\textcolor{tblGroup}{Open-weight}} \\
Search-R1-7B & \alignedscore{27.90}{} & \alignedscore{44.35}{} & \alignedscore{---}{} & \alignedscore{35.22}{} \\
WebThinker-32B-DPO & \alignedscore{48.60}{} & \alignedscore{65.80}{} & \alignedscore{---}{} & \alignedscore{46.34}{} \\
WebExplorer-8B & \alignedscore{64.80}{} & \alignedscore{85.00}{} & \alignedscore{43.81}{} & \alignedscore{47.38}{} \\
Qwen3-8B & \alignedscore{64.90}{} & \alignedscore{80.75}{} & \alignedscore{60.60}{} & \alignedscore{51.74}{} \\
\midrule[0.4pt]
\multicolumn{5}{l}{\textcolor{tblGroup}{Qwen3-4B}} \\
Initial & \alignedscore{64.49}{} & \alignedscore{81.44}{} & \alignedscore{42.80}{} & \alignedscore{50.82}{} \\
SFT & \alignedscore{\thickscoreunderline{69.37}}{\textcolor{tblGain}{(+4.88)}} & \alignedscore{83.24}{\textcolor{tblGain}{(+1.80)}} & \alignedscore{44.07}{\textcolor{tblGain}{(+1.27)}} & \alignedscore{50.56}{\textcolor{tblLoss}{(-0.26)}} \\
RL(GRPO) & \alignedscore{64.32}{\textcolor{tblLoss}{(-0.17)}} & \alignedscore{83.12}{\textcolor{tblGain}{(+1.68)}} & \alignedscore{43.80}{\textcolor{tblGain}{(+1.00)}} & \alignedscore{52.36}{\textcolor{tblGain}{(+1.54)}} \\
Rubric-RL & \alignedscore{\thickscoreunderline{69.37}}{\textcolor{tblGain}{(+4.88)}} & \alignedscore{\thickscoreunderline{84.02}}{\textcolor{tblGain}{(+2.58)}} & \alignedscore{\thickscoreunderline{45.20}}{\textcolor{tblGain}{(+2.40)}} & \alignedscore{\thickscoreunderline{54.93}}{\textcolor{tblGain}{(+4.11)}} \\
RuscaRL & \alignedscore{69.27}{\textcolor{tblGain}{(+4.78)}} & \alignedscore{81.24}{\textcolor{tblLoss}{(-0.20)}} & \alignedscore{41.92}{\textcolor{tblLoss}{(-0.88)}} & \alignedscore{53.08}{\textcolor{tblGain}{(+2.26)}} \\
\textbf{Ours} & \gainshadedscore{50.00}{\textbf{71.92}}{\textcolor{tblGain}{(+7.43)}} & \gainshadedscore{29.61}{\textbf{85.84}}{\textcolor{tblGain}{(+4.40)}} & \gainshadedscore{32.30}{\textbf{47.60}}{\textcolor{tblGain}{(+4.80)}} & \gainshadedscore{28.20}{\textbf{55.01}}{\textcolor{tblGain}{(+4.19)}} \\
\bottomrule[1pt]
\end{tabular}
\end{table}

\Needspace{5\baselineskip}
\subsection{Experimental Setup}
\label{sec:experimental_setup}
\paragraph{Benchmarks and evaluation.}
We use 487 ResearchQA questions \citep{yifei2025researchqa}, the 100-question SQAv2 test split \citep{bragg2026astabench}, GPQA-Diamond's 198 questions \citep{rein2024gpqa}, and 500 science questions from RubricHub excluded from training \citep{li2026rubrichub}. SQAv2 is closed-book and averages ingredient recall and answer precision, excluding citation metrics. Open-ended responses are graded by Gemma-4-31B-it, an open-weight judge evaluated in JudgeArena \citep{lushtaku2026judge}. A fixed 128-question RubricHub subset supports capability tracking and development monitoring. Appendix~\ref{app:evaluation} specifies scoring and generation protocols.

\paragraph{Baselines and training.}
Table~\ref{tab:main} compares the starting Qwen3-4B checkpoint, SFT, RL, Rubric-RL, RuscaRL \citep{zhou2026ruscarl}, and RSRL. SFT uses 606 question--response pairs matched to the training questions; Rubric-RL averages standardized advantages over active criteria. RL comparisons share the starting checkpoint, question schedule, and optimization rollout budget. We train Qwen3-4B \citep{yang2025qwen3} with LoRA \citep{hu2022lora} on 2,460 RubricHub science questions for 100 updates, with 32 questions per update. Training criteria are scored by a local gpt-oss-120b judge, separately from evaluation. DeepSeek-V4-Pro constructs channels and mappings, which are frozen before training. Appendices~\ref{app:channels} and~\ref{app:setup} detail construction, configurations, and baseline implementations.

\Needspace{5\baselineskip}
\subsection{Overall Performance}
\label{sec:overall_performance}
\noindent\textbf{Comparison with proprietary models.}
Against the external reference scores in Table~\ref{tab:main}, RSRL's 71.92 on ResearchQA exceeds GPT-4.1's 70.80. Its 85.84 on SQAv2 approaches Claude Opus 4.1's 86.10, highlighting the progress achieved from the 4B starting policy.

\noindent\textbf{Comparison with open-weight models.}
RSRL's ResearchQA score exceeds the listed Qwen3-8B (64.90) and WebExplorer-8B (64.80) results. Its SQAv2 score also exceeds WebExplorer-8B's 85.00 and Qwen3-8B's 80.75, placing RSRL competitively among these larger reference systems on both open-ended benchmarks.

\noindent\textbf{Comparison with post-training baselines.}
RSRL leads our Qwen3-4B comparison across all four benchmarks, ahead of SFT, RL, Rubric-RL, and RuscaRL. Gains over Rubric-RL are 2.55 points on ResearchQA, 1.82 on SQAv2, 2.40 percentage points on GPQA-Diamond, and 0.08 points on RubricHub.

\FloatBarrier
\Needspace{5\baselineskip}
\subsection{Ablation Studies}
\label{sec:ablations}
\begin{figure}[t]
\centering
\includegraphics[width=0.90\linewidth]{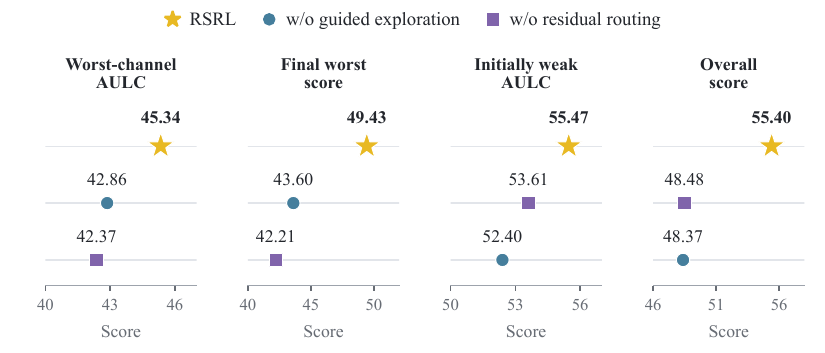}
\caption{\textbf{Effects of component removal.} Points show the four performance metrics in Table~\ref{tab:ablations}. Within each metric, methods are ordered from highest to lowest score. Colors and symbols identify methods; horizontal axes use metric-specific ranges.}
\Description{Four dot plots compare full RSRL with guided-exploration and residual-routing ablations. Full RSRL has the highest values in all four metrics.}
\label{fig:ablation_components}
\end{figure}

\noindent\textbf{Component contributions.}
We retain capability channels and remove guided exploration or residual routing in turn (Table~\ref{tab:ablations}; Figure~\ref{fig:ablation_components}). The exploration ablation uses only unguided optimization responses; the routing ablation retains guided sampling and proposal correction but uses the normalized channel-balanced baseline. Appendix~\ref{app:ablations} specifies the comparison.

\begin{table}[htbp]
\centering
\caption{\textbf{Component ablations on the 128-question development probe.} Metrics follow Appendix~\ref{app:evaluation}; scores/AULCs use 0--100 and changes use score points. Bold marks column maxima. Parentheses show differences from full RSRL; red indicates decreases.}
\label{tab:ablations}

\small
\renewcommand{\arraystretch}{1.3}
\setlength{\aboverulesep}{0pt}
\setlength{\belowrulesep}{0pt}
\setlength{\tabcolsep}{2.5pt}
\newlength{\abScoreWidth}
\newlength{\abDeltaWidth}
\settowidth{\abScoreWidth}{\textbf{88.88}}
\settowidth{\abDeltaWidth}{(-0.00)}
\newcommand{\abscore}[2]{\makebox[\abScoreWidth][r]{#1}\makebox[\abDeltaWidth][l]{#2}}
\newcommand{\abhead}[1]{\makebox[\abScoreWidth][c]{\textbf{#1}}\makebox[\abDeltaWidth]{}}
\newlength{\abSignedWidth}
\settowidth{\abSignedWidth}{$\mathbf{+1.38}$}
\newcommand{\abws}[1]{\makebox[\abSignedWidth][c]{#1}}
\begin{tabular}{lrrrr@{\hspace{9pt}}r}
\toprule[1pt]
\textbf{Method} & \abhead{W.C.} & \abhead{F.W.} & \abhead{O.W.} & \abws{\textbf{W.S.}} & \abhead{Overall} \\
\midrule[0.4pt]
\textbf{RSRL} & \abscore{\textbf{45.34}}{} & \abscore{\textbf{49.43}}{} & \abscore{\textbf{55.47}}{} & \abws{$\mathbf{+1.38}$} & \abscore{\textbf{55.40}}{} \\
\textbf{w/o guided exploration} & \abscore{42.86}{\textcolor{tblLoss}{(-2.48)}} & \abscore{43.60}{\textcolor{tblLoss}{(-5.83)}} & \abscore{52.40}{\textcolor{tblLoss}{(-3.07)}} & \abws{$-0.32$} & \abscore{48.37}{\textcolor{tblLoss}{(-7.03)}} \\
\textbf{w/o residual routing} & \abscore{42.37}{\textcolor{tblLoss}{(-2.97)}} & \abscore{42.21}{\textcolor{tblLoss}{(-7.22)}} & \abscore{53.61}{\textcolor{tblLoss}{(-1.86)}} & \abws{$-0.64$} & \abscore{48.48}{\textcolor{tblLoss}{(-6.92)}} \\
\bottomrule[1pt]
\end{tabular}
\end{table}

\noindent\textbf{Targeted exploration.}
Removing guided exploration lowers Final Worst by 5.83 points and Overall by 7.03. Worst-Strong ($\Delta$) is $-0.32$, compared with $+1.38$ for full RSRL, linking targeted exploration to stronger performance across weak and initially strong capabilities on this probe.

\noindent\textbf{Residual credit routing.}
Removing routing lowers Final Worst by 7.22 points and Overall by 6.92, with Worst-Strong ($\Delta$) at $-0.64$. This highlights the contribution of capability-specific credit allocation.

\Needspace{5\baselineskip}
\subsection{Detailed Analysis}
\label{sec:analysis}
\label{sec:capability}

\noindent
\begin{minipage}[t]{0.55\linewidth}
\vspace{0pt}
\makeatletter\def\@captype{table}\makeatother
\centering
\caption{\textbf{Capability tracking on 128 development questions.} W.C./O.W.: current-worst/initially weak channel AULC; F.W.: final worst score; W.S.: minimum strong-channel change; Overall: mean question score. Scores/AULCs use 0--100; changes use points. Bold marks maxima.}
\label{tab:capability}

\centering
\small
\setlength{\tabcolsep}{5pt}
\renewcommand{\arraystretch}{1.35}
\setlength{\aboverulesep}{0pt}
\setlength{\belowrulesep}{0pt}
\begin{tabular}{lrrrrr}
\toprule[1pt]
\textbf{Method} & \textbf{W.C.} & \textbf{F.W.} & \textbf{O.W.} & \textbf{W.S.} & \textbf{Overall} \\
\midrule[0.4pt]
Rubric-RL & 44.23 & 47.06 & 55.02 & $-0.45$ & 54.31 \\
RuscaRL & 44.99 & 45.20 & \textbf{56.08} & $-2.31$ & 54.00 \\
\textbf{RSRL} & \textbf{45.34} & \textbf{49.43} & 55.47 & $\mathbf{+1.38}$ & \textbf{55.40} \\
\bottomrule[1pt]
\end{tabular}

\end{minipage}\hfill
\begin{minipage}[t]{0.42\linewidth}
\vspace{0pt}
\makeatletter\def\@captype{figure}\makeatother
\centering
\includegraphics[width=\linewidth]{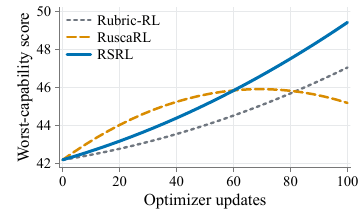}
\caption{Worst-channel scores over 100 updates: 27 checkpoints on 128 development questions.}
\Description{Worst-channel development scores for RSRL, Rubric-RL, and RuscaRL over 100 updates.}
\label{fig:capability_comparison}
\end{minipage}
\par

\noindent\textbf{Capability development.}
We track six channels at 27 checkpoints on the 128-question development probe (Figure~\ref{fig:capability_comparison}). RSRL's Final Worst exceeds Rubric-RL and RuscaRL by 2.37 and 4.23 points, with Worst-Ch. AULC gains of 1.11 and 0.35 (Table~\ref{tab:capability}). Its Worst-Strong ($\Delta$) of $+1.38$ means every initially strong channel improves; Overall gains are 1.09 and 1.40. Appendix~\ref{app:evaluation} defines the metrics.

\noindent\textbf{Guidance strength and proposal overlap.}
Offline diagnostic compares $1\times$, $3\times$, and $4\times$ guidance on 900 common-valid triggered questions.Mean target gains are 0.001747, $-0.004924$, and 0.012816 on the raw 0--1 scale, while median ESS/8 is 0.990510, 0.920261, and 0.865988. The paired $4\times-1\times$ gain is 0.011069 (95\% bootstrap interval $[-0.00402,0.02641]$). These diagnostics characterize how guidance changes both target scores and proposal overlap, motivating their joint monitoring. Appendix~\ref{app:diagnostics} reports the full statistics and the separate 50-update comparison.

\Needspace{6\baselineskip}
\section{Conclusion}
We introduced RSRL, which uses rubric structure to connect capability diagnosis, selective exploration, and response-level credit assignment. A shared capability target determines where exploration support is needed and how the resulting signals enter policy updates under the original question. Our scientific question-answering experiments show higher benchmark scores and improved weak-capability performance, alongside gains in initially strong capabilities on the development probe.
\label{sec:main_end}

\FloatBarrier
\Needspace{5\baselineskip}
\subsection*{Ethics statement}
This work uses existing question and rubric datasets and does not introduce a human-subject study. Model-generated scientific answers can be incorrect, omit relevant evidence, or express unjustified confidence. Improvements on rubric-based evaluations should not be taken as validation for scientific or other high-stakes decisions. Automated channel construction and grading may reproduce biases in the underlying models and rubric design \citep{zhang2026too}. Dataset access and any redistribution must respect the source licenses.

\Needspace{5\baselineskip}
\subsection*{Reproducibility statement}
The manuscript and appendices describe the method, training settings, evaluation protocols, and aggregate results. The source package includes numerical tables and plotting code; the appendices provide proofs of the stated proposal and routing properties. Training code, frozen checkpoints, and per-response judgments are not included. Appendix~\ref{app:record} details domain aggregation and routing reliability.

\Needspace{5\baselineskip}
\subsection*{AI use statement}
ChatGPT and Codex were used for literature search, discussion of conceptual framing and mathematical interpretation.
The experimental pipeline also uses an LLM for offline channel construction and LLM judges for
training and evaluation, as described in the paper. The human authors remain responsible for the
final claims, citations, experimental provenance, and submitted content
\bibliographystyle{ACM-Reference-Format}
\bibliography{references}

@article{gunjal2025rubrics,
 title={Rubrics as Rewards: Reinforcement Learning Beyond Verifiable Domains},
 author={Gunjal, Anisha and Wang, Anthony and Lau, Elaine and Nath, Vaskar and He, Yunzhong and Liu, Bing and Hendryx, Sean},
 journal={arXiv preprint arXiv:2507.17746}, year={2025}, url={https://arxiv.org/abs/2507.17746}}

@article{zhou2026ruscarl,
 title={Breaking the Exploration Bottleneck: Rubric-Scaffolded Reinforcement Learning for General {LLM} Reasoning},
 author={Zhou, Yang and Li, Sunzhu and Liu, Shunyu and Fang, Wenkai and Zhang, Kongcheng and Zhao, Jiale and Yang, Jingwen and Zhou, Yihe and Lv, Jianwei and Zheng, Tongya and Lu, Hengtong and Chen, Wei and Xie, Yan and Song, Mingli},
 journal={arXiv preprint arXiv:2508.16949}, year={2025}, url={https://arxiv.org/abs/2508.16949v7}}

@article{li2026rubrichub,
 title={{RubricHub}: A Comprehensive and Highly Discriminative Rubric Dataset via Automated Coarse-to-Fine Generation},
 author={Li, Sunzhu and Zhao, Jiale and Wei, Miteto and Ren, Huimin and Zhou, Yang and Yang, Jingwen and Liu, Shunyu and Zhang, Kaike and Chen, Wei},
 journal={arXiv preprint arXiv:2601.08430}, year={2026}, url={https://arxiv.org/abs/2601.08430}}

@inproceedings{shao2026drtulu,
 title={{DR Tulu}: Reinforcement Learning with Evolving Rubrics for Deep Research},
 author={Shao, Rulin and Asai, Akari and Shen, Shannon Zejiang and Ivison, Hamish and Kishore, Varsha and Zhuo, Jingming and Zhao, Xinran and Park, Molly and Finlayson, Samuel G. and Sontag, David and Murray, Tyler and Min, Sewon and Dasigi, Pradeep and Soldaini, Luca and Brahman, Faeze and Yih, Wen-tau and Wu, Tongshuang and Zettlemoyer, Luke and Kim, Yoon and Hajishirzi, Hannaneh and Koh, Pang Wei},
 booktitle={International Conference on Machine Learning}, year={2026}, url={https://arxiv.org/abs/2511.19399}}

@article{hou2026rise,
 title={{RISE-RL}: Rubric-Informed Selective Exploration for Open-Ended Reinforcement Learning},
 author={Hou, Jinkun and Liu, Zhuo and Ren, Huimin and Xin, Hongsheng and Zhou, Pan and Zhan, Kun},
 journal={arXiv preprint arXiv:2608.09123}, year={2026}, url={https://arxiv.org/abs/2608.09123}}

@article{liu2026gdpo,
 title={{GDPO}: Group reward-Decoupled Normalization Policy Optimization for Multi-reward {RL} Optimization},
 author={Liu, Shih-Yang and Dong, Xin and Lu, Ximing and Diao, Shizhe and Belcak, Peter and Liu, Mingjie and Chen, Min-Hung and Yin, Hongxu and Wang, Yu-Chiang Frank and Cheng, Kwang-Ting and Choi, Yejin and Kautz, Jan and Molchanov, Pavlo},
 journal={arXiv preprint arXiv:2601.05242}, year={2026}, url={https://arxiv.org/abs/2601.05242}}

@article{liu2026gd2po,
 title={{GD}$^2${PO}: Mitigating Multi-Reward Conflicts via Group-Dynamic reward-Decoupled Policy Optimization},
 author={Liu, Haotian and Liu, Yihao and Ni, Jingwei and Huang, Siyuan and Liu, Xinpeng and Cheng, Pengyu and Song, Jiajun and Ding, Ruijin and Li, Junfeng and Yu, Zhechao and Zhou, Mengyu and Xu, Hongteng and Jiang, Xiaoxi and Jiang, Guanjun},
 journal={arXiv preprint arXiv:2606.16771}, year={2026}, url={https://arxiv.org/abs/2606.16771}}

@inproceedings{graves2017curriculum,
 title={Automated Curriculum Learning for Neural Networks},
 author={Graves, Alex and Bellemare, Marc G. and Menick, Jacob and Munos, R{\'e}mi and Kavukcuoglu, Koray},
 booktitle={Proceedings of the 34th International Conference on Machine Learning}, volume={70}, pages={1311--1320}, year={2017}, publisher={PMLR}, url={https://proceedings.mlr.press/v70/graves17a.html}}

@inproceedings{li2023contrastive,
 title={Contrastive Decoding: Open-ended Text Generation as Optimization},
 author={Li, Xiang Lisa and Holtzman, Ari and Fried, Daniel and Liang, Percy and Eisner, Jason and Hashimoto, Tatsunori and Zettlemoyer, Luke and Lewis, Mike},
 booktitle={Proceedings of the 61st Annual Meeting of the Association for Computational Linguistics (Volume 1: Long Papers)},
 year={2023}, pages={12286--12312}, publisher={Association for Computational Linguistics}, doi={10.18653/v1/2023.acl-long.687}, url={https://aclanthology.org/2023.acl-long.687/}}

@inproceedings{pei2023preadd,
 title={{PREADD}: Prefix-Adaptive Decoding for Controlled Text Generation},
 author={Pei, Jonathan and Yang, Kevin and Klein, Dan},
 booktitle={Findings of the Association for Computational Linguistics: ACL 2023}, year={2023}, pages={10018--10037},
 publisher={Association for Computational Linguistics}, doi={10.18653/v1/2023.findings-acl.636}, url={https://aclanthology.org/2023.findings-acl.636/}}

@inproceedings{veach1995optimally,
 title={Optimally Combining Sampling Techniques for {Monte Carlo} Rendering},
 author={Veach, Eric and Guibas, Leonidas J.}, booktitle={Proceedings of SIGGRAPH}, year={1995}, pages={419--428}, url={https://graphics.stanford.edu/papers/combine/}}

@article{yang2025qwen3,
 title={{Qwen3} Technical Report}, author={Yang, An and Li, Anfeng and Yang, Baosong and Zhang, Beichen and Hui, Binyuan and Zheng, Bo and Yu, Bowen and others},
 journal={arXiv preprint arXiv:2505.09388}, year={2025}, url={https://arxiv.org/abs/2505.09388}}

@inproceedings{hu2022lora,
 title={{LoRA}: Low-Rank Adaptation of Large Language Models},
 author={Hu, Edward J. and Shen, Yelong and Wallis, Phillip and Allen-Zhu, Zeyuan and Li, Yuanzhi and Wang, Shean and Wang, Lu and Chen, Weizhu},
 booktitle={International Conference on Learning Representations}, year={2022}, url={https://arxiv.org/abs/2106.09685}}

@article{yifei2025researchqa,
 title={{ResearchQA}: Evaluating Scholarly Question Answering at Scale Across 75 Fields with Survey-Mined Questions and Rubrics},
 author={Yifei, Li S. and Chang, Allen and Malaviya, Chaitanya and Yatskar, Mark},
 journal={arXiv preprint arXiv:2509.00496}, year={2025}, url={https://arxiv.org/abs/2509.00496}}

@inproceedings{rein2024gpqa,
 title={{GPQA}: A Graduate-Level Google-Proof {Q\&A} Benchmark},
 author={Rein, David and Hou, Betty Li and Stickland, Asa Cooper and Petty, Jackson and Pang, Richard Yuanzhe and Dirani, Julien and Michael, Julian and Bowman, Samuel R.},
 booktitle={First Conference on Language Modeling}, year={2024}, url={https://arxiv.org/abs/2311.12022}}

@inproceedings{bragg2026astabench,
 title={{AstaBench}: Rigorous Benchmarking of {AI} Agents with a Scientific Research Suite},
 author={Bragg, Jonathan and D'Arcy, Mike and Balepur, Nishant and Bareket, Dan and Dalvi, Bhavana and Feldman, Sergey and Haddad, Dany and Hwang, Jena D. and Jansen, Peter and Kishore, Varsha and others},
 booktitle={International Conference on Learning Representations}, year={2026}, url={https://arxiv.org/abs/2510.21652}}

@article{wang2026smopd,
 title={{SMOPD}: Multi-Reward Reinforcement Learning via Specialize-and-Merge Online Policy Distillation},
 author={Wang, Wen and Bao, Jiahua and Yongsiqi, Tu and Liu, Yihao and Zhou, Haotian and Ma, Haoxuan and Zhou, Mengyu and Fan, Wenkui and He, Junwei and Jiang, Xiaoxi and Jiang, Guanjun},
 journal={arXiv preprint arXiv:2608.03092}, year={2026}, url={https://arxiv.org/abs/2608.03092v2}}

@article{lushtaku2026judge,
 title={{JudgeArena}: A Unified Framework for Reproducible {LLM}-Judge Evaluation},
 author={Lushtaku, Erlis and Kargi, Bora and Elganzory, Ali and Ferreira, Fabio and Salamanca, Alejandro R. and Kreutzer, Julia and Salinas, David},
 journal={arXiv preprint arXiv:2608.02620}, year={2026},
 url={https://arxiv.org/abs/2608.02620}}

@inproceedings{kim2024prometheus,
 title={Prometheus: Inducing Fine-Grained Evaluation Capability in Language Models},
 author={Kim, Seungone and Shin, Jamin and Cho, Yejin and Jang, Joel and Longpre, Shayne and Lee, Hwaran and Yun, Sangdoo and Shin, Seongjin and Kim, Sungdong and Thorne, James and Seo, Minjoon},
 booktitle={The Twelfth International Conference on Learning Representations}, year={2024},
 url={https://openreview.net/forum?id=8euJaTveKw}}

@article{arora2025healthbench,
 title={{HealthBench}: Evaluating Large Language Models Towards Improved Human Health},
 author={Arora, Rahul K. and Wei, Jason and Hicks, Rebecca Soskin and Bowman, Preston and Qui{\~n}onero-Candela, Joaquin and Tsimpourlas, Foivos and Sharman, Michael and Shah, Meghan and Vallone, Andrea and Beutel, Alex and Heidecke, Johannes and Singhal, Karan},
 journal={arXiv preprint arXiv:2505.08775}, year={2025}, url={https://arxiv.org/abs/2505.08775}}

@article{gupta2026craft,
 title={{CRAFT}: Clustering Rubrics to Diagnose Weak {LLM} Capabilities and Generate Targeted Fine-Tuning Data},
 author={Gupta, Vipul and Wang, Zihao and Dumitru, Razvan-Gabriel and Rezaei, MohammadHossein and Sabharwal, Aakash and He, Yunzhong},
 journal={arXiv preprint arXiv:2607.16122}, year={2026}, url={https://arxiv.org/abs/2607.16122}}

@inproceedings{zhang-etal-2026-semantic,
    title = "Semantic-Aware Logical Reasoning via a Semiotic Framework",
    author = "Zhang, Yunyao  and
      Zhang, Xinglang  and
      Sheng, Junxi  and
      Li, Wenbing  and
      Yu, Junqing  and
      Chen, Yi-Ping Phoebe  and
      Yang, Wei  and
      Song, Zikai",
    editor = "Liakata, Maria  and
      Moreira, Viviane P.  and
      Zhang, Jiajun  and
      Jurgens, David",
    booktitle = "Proceedings of the 64th Annual Meeting of the {A}ssociation for {C}omputational {L}inguistics (Volume 1: Long Papers)",
    month = jul,
    year = "2026",
    address = "San Diego, California, United States",
    publisher = "Association for Computational Linguistics",
    url = "https://aclanthology.org/2026.acl-long.835/",
    doi = "10.18653/v1/2026.acl-long.835",
    pages = "18349--18374",
    ISBN = "979-8-89176-390-6"
}

@inproceedings{zhang-etal-2026-logical,
    title = "Logical Phase Transitions: Understanding Collapse in {LLM} Logical Reasoning",
    author = "Zhang, Xinglang  and
      Zhang, Yunyao  and
      Chen, ZeLiang  and
      Yu, Junqing  and
      Yang, Wei  and
      Song, Zikai",
    editor = "Liakata, Maria  and
      Moreira, Viviane P.  and
      Zhang, Jiajun  and
      Jurgens, David",
    booktitle = "Proceedings of the 64th Annual Meeting of the {A}ssociation for {C}omputational {L}inguistics (Volume 1: Long Papers)",
    month = jul,
    year = "2026",
    address = "San Diego, California, United States",
    publisher = "Association for Computational Linguistics",
    url = "https://aclanthology.org/2026.acl-long.858/",
    doi = "10.18653/v1/2026.acl-long.858",
    pages = "18836--18860",
    ISBN = "979-8-89176-390-6"
}

@article{zhang2026coupling,
  title={Coupling Macro Dynamics and Micro States for Long-Horizon Social Simulation},
  author={Zhang, Yunyao and Ai, Yihao and Ying, Zuocheng and Mi, Qirui and Yu, Junqing and Yang, Wei and Song, Zikai},
  journal={arXiv preprint arXiv:2604.05516},
  year={2026}
}

@misc{zhang2026intervensiminterventionawaresocialnetwork,
      title={IntervenSim: Intervention-Aware Social Network Simulation for Opinion Dynamics}, 
      author={Yunyao Zhang and Zuocheng Ying and Xinglang Zhang and Junqing Yu and Peng Fang and Xu Chen and Wei Yang and Zikai Song},
      year={2026},
      eprint={2604.06600},
      archivePrefix={arXiv},
      primaryClass={cs.SI},
      url={https://arxiv.org/abs/2604.06600}, 
}

@misc{wang2026seeingwiderjointspatiotemporal,
      title={Seeing Further and Wider: Joint Spatio-Temporal Enlargement for Micro-Video Popularity Prediction}, 
      author={Dali Wang and Yunyao Zhang and Junqing Yu and Yi-Ping Phoebe Chen and Chen Xu and Zikai Song},
      year={2026},
      eprint={2604.20311},
      archivePrefix={arXiv},
      primaryClass={cs.MM},
      url={https://arxiv.org/abs/2604.20311}, 
}

@article{zheng2026surfacing,
  title={Surfacing the Unsaid: CUE-Bench for Affective Stance in Chinese Discourse},
  author={Zheng, Zhenyan and Zhang, Yunyao and Sheng, Junxi and Yu, Junqing and Song, Zikai},
  journal={arXiv preprint arXiv:2608.10810},
  year={2026}
}

@misc{song2026socialintelligence,
  title        = {Social Intelligence Modeling: A Comprehensive Survey from Social Perception to Social Simulation},
  author       = {Song, Zikai and Li, Xiajie and Zhang, Yunyao and Zhang, Xinglang and Yang, Wei and Yu, Junqing},
  howpublished = {ResearchGate preprint},
  year         = {2026},
  note         = {Preprint available on ResearchGate},
  doi          = {10.13140/RG.2.2.21157.87528},
  url          = {https://www.researchgate.net/publication/411123257_Social_Intelligence_Modeling_A_Comprehensive_Survey_from_Social_Perception_to_Social_Simulation}
}

@article{zhang2026too,
  title={Too Good to Be Real? Diagnosing and Reducing the Gap Between AI Preference and Real User Engagement},
  author={Zhang, Xinglang and Xiang, Yuanmeng and Zhang, Yunyao and Chen, Zeliang and Yu, Junqing and Song, Zikai},
  journal={arXiv preprint arXiv:2609.18282},
  year={2026}
}

@inproceedings{li2026lora,
  title={Lora-mixer: Coordinate modular lora experts through serial attention routing},
  author={Li, Wenbing and Song, Zikai and Zhou, Hang and Yu, Junqing and Zhang, Yunyao and Yang, Wei},
  booktitle={International Conference on Learning Representations},
  volume={2026},
  pages={14694--14716},
  year={2026}
}

@inproceedings{song2025temporal,
  title={Temporal Coherent Object Flow for Multi-Object Tracking},
  author={Song, Zikai and Luo, Run and Ma, Lintao and Tang, Ying and Chen, Yi-Ping Phoebe and Yu, Junqing and Yang, Wei},
  booktitle={Proceedings of the AAAI Conference on Artificial Intelligence},
  volume={39},
  number={7},
  pages={6978--6986},
  year={2025},
  doi={10.1609/aaai.v39i7.32749},
  url={https://ojs.aaai.org/index.php/AAAI/article/view/32749}
}

@article{song2026hypergraph,
  title={Hypergraph-State Collaborative Reasoning for Multi-Object Tracking},
  author={Song, Zikai and Yu, Junqing and Chen, Yi-Ping Phoebe and Yang, Wei and Wang, Xinchao},
  journal={arXiv preprint arXiv:2604.12665},
  year={2026},
  url={https://arxiv.org/abs/2604.12665}
}

@inproceedings{song2023compact,
  title={Compact Transformer Tracker with Correlative Masked Modeling},
  author={Song, Zikai and Luo, Run and Yu, Junqing and Chen, Yi-Ping Phoebe and Yang, Wei},
  booktitle={Proceedings of the AAAI Conference on Artificial Intelligence},
  volume={37},
  number={2},
  pages={2321--2329},
  year={2023},
  doi={10.1609/aaai.v37i2.25327},
  url={https://ojs.aaai.org/index.php/AAAI/article/view/25327}
}
\FloatBarrier
\appendix
\FloatBarrier
\Needspace{6\baselineskip}
\section{Capability construction and diagnostic scores}
\label{app:channels}
\Needspace{5\baselineskip}
\subsection{Frozen domain and local channels}
We construct the domain codebook from 64 questions sampled from training, calibration, and development records, excluding evaluation and reserve splits. Selection uses a fixed hash ordering. The construction model receives the questions and their source rubrics. It proposes three to eight recurring capabilities, after which coverage checks and consolidation yield six science channels: quantitative and formal derivation; precise technical communication and structured setup; mechanistic and causal explanation; experimental, computational, and intervention design; comparison and synthesis across alternatives; and assessment of evidence, assumptions, and scope.

Each question's source criteria are grouped into two to five local channels, with at least two criteria per channel. Canonical source identifiers prevent aliases from receiving duplicate assignments. Programmatic checks enforce coverage and unique assignment; model-based review checks grouping coherence. The frozen training set contains 2,460 questions after construction checks. These local channels retain question-specific requirements while the domain codebook supplies shared categories.

\Needspace{5\baselineskip}
\subsection{Mapping local channels to the domain codebook}
Construction and mapping use the API identifier \texttt{deepseek-v4-pro}, with temperature zero, and JSON output. The mapping judge receives the question, local channel definition, and domain codebook. It returns a domain identifier, applicability decision, confidence score, and rationale.

For each local channel, two separate requests to the same model produce mappings $m_q^A(c),m_q^B(c)$ and confidence scores $p_{q,c}^A,p_{q,c}^B$. We accept a mapping when both requests judge it applicable and
\[
m_q^A(c)=m_q^B(c)\ne\bot,\qquad \min(p_{q,c}^A,p_{q,c}^B)\geq0.8.
\]
The accepted confidence is $m_{q,c}^{\mathrm{conf}}=\min(p_{q,c}^A,p_{q,c}^B)$. Otherwise, the channel remains local-only, with $m_q(c)=\bot$ and $m_{q,c}^{\mathrm{conf}}=0$. These are two calls to one model, and their self-reported scores are not calibrated probabilities. Definitions, assignments, and mapping confidences are frozen before policy training. Appendix~\ref{app:record} describes their use in diagnosis.

\Needspace{5\baselineskip}
\subsection{Rubric banks and valid observations}
Within each channel, diagnostic and training banks use the same set of at most five canonical criteria, selected by descending absolute source points with fixed hash tie-breaking. The evaluation bank retains the source criteria required by its evaluation protocol. Evaluation criteria are not supplied as generation guidance.

Let $s_{jr}\in[0,1]$ be the polarity-aligned diagnostic score for response $j$ and criterion $r$, and let $v_{jr}$ indicate validity. Scores are normalized by the criterion's predefined range. A valid zero remains an observation. For $N_r^D=\sum_j v_{jr}>0$, define
\[
\mu_r^D=\frac{\sum_j v_{jr}s_{jr}}{N_r^D},\qquad
\mathcal F_{q,c}=\{r\in\mathcal R_{q,c}^D:N_r^D>0\}.
\]
Using frozen criterion reliability $u_r$, the local score and reliability are
\[
S_{q,c}=\frac{\sum_{r\in\mathcal F_{q,c}}u_r\mu_r^D}{\sum_{r\in\mathcal F_{q,c}}u_r},\qquad
r_{q,c}=\frac{1}{|\mathcal F_{q,c}|}\sum_{r\in\mathcal F_{q,c}}u_r.
\]
Thus, a wholly missing criterion is omitted from both numerator and denominator. The science banks use $u_r=1$, giving equal weight to available criterion means. If $\mathcal F_{q,c}$ is empty, the local score is unavailable and $r_{q,c}=0$; diagnosis then follows the missing-evidence rules in Section~\ref{sec:capability_structuring}.

The count used for historical pooling is
\[
n_{q,c}=\sum_{j=1}^{4}\mathbf1\bigl[\exists r\in\mathcal R_{q,c}^D:v_{jr}=1\bigr].
\]
One valid criterion suffices to count a response once. This count differs from the stricter criterion eligibility tests for guided exploration and optimization advantages below.

\FloatBarrier
\Needspace{6\baselineskip}
\section{Training procedure and estimator details}
\label{app:algorithm}
\Needspace{5\baselineskip}
\subsection{One policy update}
An unguided diagnostic pass initializes the domain state without gradient updates. Each subsequent update uses the following order:
\begin{enumerate}[leftmargin=*,itemsep=2pt]
\item Snapshot the policy and committed domain state. Generate four unguided diagnostic responses for each scheduled question.
\item Score the diagnostic criteria, pool local and historical evidence, and select the target capability. Hold this target fixed for the question's update.
\item Apply the exploration support test. Generate four unguided and four guided optimization responses if active, or eight unguided responses otherwise. Evaluate the proposal densities and weights.
\item Score the optimization responses, construct the baseline and routed candidate, and apply the guard after final normalization.
\item Accumulate the clean-context loss over the question batch, clip the accumulated gradient, and take one optimizer step.
\item Aggregate the unguided diagnostic observations within questions and then equally across questions. Commit the domain update for use at the next step.
\end{enumerate}
All questions in an update use the same historical snapshot. Guided responses do not update the domain state.

\Needspace{5\baselineskip}
\subsection{Support detection and exploration strength}
\label{app:ctse}
For the selected channel, let $\mathcal E=\{r:N_r^D\geq4\}$. Since diagnosis uses four responses, only criteria with all four valid judgments enter the support test. For each eligible criterion, compute its population variance $V_r$ and positive-response fraction $p_r$, using score $0.5$ as the positive threshold. The \textsc{Zero} regime requires all eligible scores to be zero. The \textsc{Low} regime requires at least one $V_r>10^{-4}$ and $|\mathcal E|^{-1}\sum_{r\in\mathcal E}p_r\leq0.25$. An empty eligible set or any other support state leaves sampling unguided.

The coefficient in Section~\ref{sec:mis} is
\begin{equation}
\eta_q=\frac{\eta_{\max}}{\sqrt{H_{\max}}}\clip\!\left(\frac{h_q^*-0.05}{0.30-0.05},0,1\right)g_{\mathrm{support}},
\end{equation}
where $g_{\mathrm{support}}$ is one for \textsc{Zero}, one-half for \textsc{Low}, and zero otherwise. Here $H_{\max}=3072$ is the response cap and $\eta_{\max}=1$, so the effective coefficient is capped at $1/\sqrt{3072}$. The strength therefore vanishes for deficits at most $0.05$ and saturates at $0.30$.

The target prompt contains the selected capability's description and associated training criteria, together with an instruction not to disclose the internal guidance. The control prompt supplies nondirectional guidance. The three distributions are evaluated at the same response prefix; the optimization loss is always conditioned on the original question.

\Needspace{5\baselineskip}
\subsection{Criterion advantages and final normalization}
\label{app:credit}
Let $z_{ir}$ be an optimization score oriented so that higher is better, with validity $v_{ir}$. For a criterion with $N_r=\sum_i v_{ir}>0$, compute unweighted moments
\begin{equation}
\mu_r=\frac{\sum_i v_{ir}z_{ir}}{N_r},\qquad
\sigma_r^2=\frac{\sum_i v_{ir}(z_{ir}-\mu_r)^2}{N_r}.
\end{equation}
A criterion is active if it is canonical, has frozen reliability at least $0.8$, has at least $\max(4,\lceil0.75G\rceil)=6$ valid observations, and has standard deviation greater than $0.05$ times its predefined score range. Valid active entries use $a_{ir}=\clip((z_{ir}-\mu_r)/\sigma_r,-5,5)$; all other entries are zero. Fixed within-channel weights are not redistributed when criteria become inactive.

The deficit schedule and reference scale in Section~\ref{sec:rar} are
\begin{align}
f(h)&=\lambda_{\max}\left[\clip\!\left(\frac{h-h_R}{1-h_R},0,1\right)\right]^p,\\
\|C\|_w^2&=\sum_i w_i\frac{1}{|\mathcal C_q|}\sum_c C_{ic}^2,\qquad
r_{\mathrm{ref}}=\max\bigl(\|B\|_w,\min(1,\|C\|_w),\varepsilon\bigr).
\end{align}
We set $R'=R$ when $\langle B,B\rangle_w\leq\varepsilon^2$ and omit the correction when $\|R'\|_w\leq10^{-6}$. Single-channel questions and zero routing strength use the baseline.

The final normalization is
\begin{align}
\widehat X_i&=\clip\!\left(\frac{X_i}{\max(\|X\|_w,\varepsilon)},-5,5\right),\qquad
V_i=\widehat X_i-\sum_j w_j\widehat X_j,\\
Z_i&=\frac{V_i}{\max(\|V\|_w,\varepsilon)},\qquad
\mathcal N_w(X)_i=Z_i\min\!\left(1,\frac5{\max_j|Z_j|}\right).
\end{align}
The last factor is one when $Z=0$. Both candidate and baseline use this order of operations. No further normalization follows the guard decision. Table~\ref{tab:parameters} lists the fixed parameters.

\Needspace{5\baselineskip}
\subsection{Sequence densities and the implemented loss}
\label{app:update}
Training uses temperature one, top-$p$ one, and no top-$k$ truncation. The target--control contrast uses full softmax probabilities. Sequence log densities include all realized tokens, including sampled EOS, but exclude padding and any unobserved EOS at the length cap. Teacher-forced replay evaluates the guided density of unguided responses, and log-sum-exp evaluates the mixture density.

We monitor $\mathrm{ESS}=(\sum_i\omega_i)^2/\sum_i\omega_i^2$ and guided mass $\mathrm{Share}_g=\sum_{i\ \mathrm{guided}}w_i$. Valid branch densities and $\mathrm{ESS}/G\geq0.25$ are required; failure raises an error without modifying the weights. Small guided weights are not floored.

Write $\ell_{qit}(\theta)=\log\pi_\theta(y_{qit}\mid x_q,y_{qi,<t})$. With all sampled quantities held fixed, the implemented loss is
\begin{equation}
\mathcal L_{\mathrm{impl}}(\theta)=\frac1{|\mathcal B|G}\sum_{q,i}\omega_{qi}
\left[-\frac1{L_{qi}}\sum_t\min(r_{qit}A_{qi},\clip(r_{qit},0.8,1.2)A_{qi})+\beta_{\mathrm{KL}}\bar k_{qi}(\theta)\right],
\end{equation}
\begin{align}
r_{qit}&=\exp\bigl(\ell_{qit}(\theta)-\sg(\ell_{qit}(\theta))\bigr),\\
\bar k_{qi}(\theta)&=\frac1{L_{qi}}\sum_t\left(e^{d_{qit}(\theta)}-d_{qit}(\theta)-1\right),\nonumber\\
d_{qit}(\theta)&=\log\pi_{\mathrm{ref}}(y_{qit}\mid x_q,y_{qi,<t})-\ell_{qit}(\theta).
\end{align}
Here $\sg$ denotes stop-gradient and $\pi_{\mathrm{ref}}$ is the frozen starting policy. During the single pre-update pass, $r_{qit}=1$ with nonzero derivative, so clipping is inactive and the policy term has the local gradient stated in Equation~\eqref{eq:loss}. The reference penalty remains differentiable. Raw proposal weights multiply both terms. This equality describes the specified one-pass update.

\FloatBarrier
\Needspace{6\baselineskip}
\section{Properties of exploration and routing}
\label{app:properties}
\Needspace{5\baselineskip}
\subsection{Proposal support and importance correction}
At a fixed prefix, the bound $|\delta(v)|\leq b$ places both $\exp(\eta\delta(v))$ and its normalizer in $[e^{-\eta b},e^{\eta b}]$. Hence
\begin{equation}
e^{-2\eta b}\leq\frac{q_\eta(v)}{p_0(v)}\leq e^{2\eta b}\quad\text{for }p_0(v)>0.
\end{equation}
Finite perturbations preserve probability support. The diagnostic \textsc{Zero} label concerns observed scores, not zero probability under the policy.

Conditioning on diagnosis fixes $P_0$ and $Q_\eta$. For an integrable function $f$ independent of the sampled optimization group, the equal-allocation balance estimator satisfies
\begin{equation}
\frac12\E_{P_0}\!\left[\frac{P_0}{M}f\right]+\frac12\E_{Q_\eta}\!\left[\frac{P_0}{M}f\right]
=\int M(y)\frac{P_0(y)}{M(y)}f(y)\,dy=\E_{P_0}[f].
\end{equation}
Also, $M\geq P_0/2$ gives the ratio bound in Section~\ref{sec:mis}. This identity applies to a fixed integrand; criterion normalization and routing depend on the sampled group, so it does not establish unbiasedness of the complete gradient estimator.

For fixed $p_0$ and $\delta$, let $Z(\eta)=\E_{p_0}e^{\eta\delta}$. Then
\begin{equation}
D_{\mathrm{KL}}(p_0\Vert q_\eta)=\log Z(\eta)-\eta\E_{p_0}\delta,\qquad
\frac{d^2}{d\eta^2}D_{\mathrm{KL}}(p_0\Vert q_\eta)=\operatorname{Var}_{q_\eta}(\delta)\geq0.
\end{equation}
The first derivative is zero at $\eta=0$, so this divergence is nondecreasing for $\eta\geq0$ at the fixed prefix. This is the setting used for the replay diagnostic in Appendix~\ref{app:diagnostics}.

\Needspace{5\baselineskip}
\subsection{Normalization and guard guarantees}
\label{app:guard}
Assume finite channel advantages, $w_i\geq0$, $\sum_i w_i=1$, and the numerical conventions in Appendix~B.3. Partial projection satisfies $\langle R',B\rangle_w=(1-\xi)\langle R,B\rangle_w$ on its nondegenerate branch; it attenuates the baseline-parallel component. When applied, the correction has $\|H-B\|_w=\lambda_qr_{\mathrm{ref}}$.

\noindent\textbf{Proposition C.1.} \textit{For every finite $X$, $\mathcal N_w(X)$ has weighted mean zero, weighted norm at most one, and coordinate magnitude at most five. It preserves weak coordinate ordering.}

\noindent\textit{Proof.} The first positive common scaling and nondecreasing clipping preserve order. Subtracting a common weighted mean centers the vector without changing pairwise differences. Division by $\max(\|V\|_w,\varepsilon)$ gives weighted norm at most one. The final common factor is positive and at most one, preserves centering and order, and enforces the coordinate bound.\hfill$\square$

\noindent\textbf{Proposition C.2.} \textit{The advantage $A$ selected in Equation~\eqref{eq:selected} has the preceding normalization properties. For every $(i,j)\in\mathcal P$, $A_i\geq A_j$. In addition,}
\[
\langle T,A-A^0\rangle_w\geq0,\qquad r_{\mathrm{ref}}\|A-A^0\|_w\leq\|R'\|_w.
\]
\textit{The first inequality is strict whenever the candidate is accepted.}

\noindent\textit{Proof.} On acceptance, the ordering and both inequalities are explicit guard conditions. On rejection, $A=A^0$: its ordering follows that of $B$, and its change is zero. Either selected vector has the normalization properties, and no subsequent transformation is applied.\hfill$\square$

A constant target has no strict target preference. The same opportunity set is empty for $T=aB+b\mathbf1$ with $a>0$. Both cases return $A^0$. Moreover, if $\|R'\|_w/r_{\mathrm{ref}}\to0$, the weighted norm of the final credit change tends to zero. 

\Needspace{5\baselineskip}
\subsection{Boundary and acceptance examples}
Let eight responses have equal weights and define orthogonal vectors with unit weighted norm
\[
u=(-1,-1,-1,-1,1,1,1,1),\qquad v=(-1,-1,1,1,-1,-1,1,1).
\]
For $C_1=0$, $C_2=u$, and $C_3=-0.8u+0.2v$, selecting channel one gives $T=0$, $B=(0.2u+0.2v)/3$, $\|B\|_w=0.094281$, and $r_{\mathrm{ref}}=0.748331$. With $\xi=0.5$ and $\lambda_q=0.45$, the unguarded candidate is $H=-2.571764B$. The guard returns $A^0$ because there is no target opportunity. More generally, for $T=aB$ with $0\leq a<1$ and an applied correction,
\[
H=\left(1-\frac{\lambda_qr_{\mathrm{ref}}}{\|B\|_w}\right)B.
\]
This exposes the aligned-target boundary even when the target varies across responses.

For an accepted case, take $C_1=u$, $C_2=v$, and $T=u$. Then $B=(u+v)/2$, $R'=(u-v)/2$, and $r_{\mathrm{ref}}=1$. At $\lambda_q=0.45$,
\[
A^0=(u+v)/\sqrt2,\qquad A^1=0.9761921u+0.2169079v.
\]
There is a target preference within a baseline tie, shared orderings are preserved, and $\langle T,\Delta A\rangle_w=0.2690853>0$. The weighted norm of the change is $0.5591975<1/\sqrt2$, so all four conditions hold. These channel vectors can be formed from distinct binary criteria with repeated score patterns; the examples describe the rule's behavior, not experimental frequencies.

\FloatBarrier
\Needspace{6\baselineskip}
\section{Experimental settings and evaluation}
\label{app:setup}
\Needspace{5\baselineskip}
\subsection{Evaluation protocols and capability metrics}
\label{app:evaluation}
Evaluation responses use the original question without rubric or channel guidance. Open-ended scoring uses \texttt{Gemma-4-31B-it} with thinking enabled, a default context of 16,384 tokens, and an 8,192-token completion ceiling, separately from the training judge.

\paragraph{ResearchQA.}
The 487 evaluation questions, containing 3,633 source criteria, are disjoint from the 2,460 training questions. Source-weighted criterion scores are aggregated within each question and averaged equally over questions. Missing or invalid judgments receive zero. Generation uses temperature $0.6$, top-$p$ $0.95$, top-$k$ $20$, an initial response cap of 16,384, and context length 40,960. The evaluation protocol includes repetition and length repair.

\paragraph{SQAv2.}
We use the 100-question SQA-CS-V2 test split with 1,094 source ingredients and ASTA/astabench 0.3.1 scoring functions. The reported content score is $100(\mathrm{IR}+\mathrm{AP})/2$, where ingredient recall and answer precision are on a 0--1 scale. This closed-book evaluation excludes citation metrics.

\paragraph{GPQA-Diamond.}
Accuracy is averaged across four independent evaluation runs on 198 questions. Generation uses temperature $0.7$, top-$p$ $0.8$, top-$k$ $20$, a 32,768-token response cap, and context length 40,960. Format completion handles answers without a parseable option without consulting the answer key; remaining unparseable answers count as incorrect. These are inference repetitions of each trained model.

\paragraph{RubricHub and capability tracking.}
The 500-question science evaluation set is excluded from policy training. Its evaluator averages valid criterion judgments within a question, omitting invalid judgments, and then averages across questions. The fixed 128-question development probe is a subset of this set. It is evaluated at checkpoints $0,2,4,8,12,\ldots,100$, giving 27 observations.

Let $S_z(s)$ average the scores of questions applicable to capability $z$ at checkpoint $s$, and $W(s)=\min_z S_z(s)$. For $0=s_0<\cdots<s_m=S$, normalized AULC is
\begin{equation}
\mathrm{AULC}_S(W)=\frac1S\sum_{j=1}^m(s_j-s_{j-1})\frac{W(s_j)+W(s_{j-1})}{2}.
\end{equation}
The main comparison uses $S=100$. Orig.-Weak AULC replaces $W$ with the score of the fixed initially weak capability, assessment of evidence, assumptions, and scope. The prespecified strong set comprises technical communication, mechanistic explanation, and quantitative derivation. These identities come from the shared initial training diagnosis. The retention metric is
\[
\Delta_{\mathrm{strong}}=\min_{z\in\mathcal Z_{\mathrm{strong}}}\bigl[S_z(100)-S_z(0)\bigr].
\]
Overall is the mean question-level response score, not a mean over the six capabilities.

\Needspace{5\baselineskip}
\subsection{Training configuration}
\begin{table}[t]
\centering
\caption{Training configuration. Response budgets are per question. The lower block records additional RuscaRL settings.}
\label{tab:training}

\setlength{\tblValueWidth}{0pt}
\settowidth{\tblMeasureWidth}{Value}\ifdim\tblMeasureWidth>\tblValueWidth\setlength{\tblValueWidth}{\tblMeasureWidth}\fi
\settowidth{\tblMeasureWidth}{Qwen3-4B / 2,460 science questions}\ifdim\tblMeasureWidth>\tblValueWidth\setlength{\tblValueWidth}{\tblMeasureWidth}\fi
\settowidth{\tblMeasureWidth}{100 / 32}\ifdim\tblMeasureWidth>\tblValueWidth\setlength{\tblValueWidth}{\tblMeasureWidth}\fi
\settowidth{\tblMeasureWidth}{4 / 8}\ifdim\tblMeasureWidth>\tblValueWidth\setlength{\tblValueWidth}{\tblMeasureWidth}\fi
\settowidth{\tblMeasureWidth}{4 unguided + 4 guided}\ifdim\tblMeasureWidth>\tblValueWidth\setlength{\tblValueWidth}{\tblMeasureWidth}\fi
\settowidth{\tblMeasureWidth}{3,072 tokens / non-thinking}\ifdim\tblMeasureWidth>\tblValueWidth\setlength{\tblValueWidth}{\tblMeasureWidth}\fi
\settowidth{\tblMeasureWidth}{32 / 32 / 0}\ifdim\tblMeasureWidth>\tblValueWidth\setlength{\tblValueWidth}{\tblMeasureWidth}\fi
\settowidth{\tblMeasureWidth}{All linear layers / none}\ifdim\tblMeasureWidth>\tblValueWidth\setlength{\tblValueWidth}{\tblMeasureWidth}\fi
\settowidth{\tblMeasureWidth}{$10^{-5}$ / $10^{-3}$}\ifdim\tblMeasureWidth>\tblValueWidth\setlength{\tblValueWidth}{\tblMeasureWidth}\fi
\settowidth{\tblMeasureWidth}{4 / 1.0}\ifdim\tblMeasureWidth>\tblValueWidth\setlength{\tblValueWidth}{\tblMeasureWidth}\fi
\settowidth{\tblMeasureWidth}{\texttt{gpt-oss-120b}, medium reasoning}\ifdim\tblMeasureWidth>\tblValueWidth\setlength{\tblValueWidth}{\tblMeasureWidth}\fi
\settowidth{\tblMeasureWidth}{8,192 tokens}\ifdim\tblMeasureWidth>\tblValueWidth\setlength{\tblValueWidth}{\tblMeasureWidth}\fi
\settowidth{\tblMeasureWidth}{Sigmoid ($\alpha=125$, $t_0=0.2$) / linear}\ifdim\tblMeasureWidth>\tblValueWidth\setlength{\tblValueWidth}{\tblMeasureWidth}\fi
\settowidth{\tblMeasureWidth}{Scaffolded / original question}\ifdim\tblMeasureWidth>\tblValueWidth\setlength{\tblValueWidth}{\tblMeasureWidth}\fi
\settowidth{\tblMeasureWidth}{0, 0.5, 1}\ifdim\tblMeasureWidth>\tblValueWidth\setlength{\tblValueWidth}{\tblMeasureWidth}\fi
\settowidth{\tblMeasureWidth}{Temperature 1; top-$p$ 1; no top-$k$}\ifdim\tblMeasureWidth>\tblValueWidth\setlength{\tblValueWidth}{\tblMeasureWidth}\fi
\settowidth{\tblMeasureWidth}{AdamW / 0.2}\ifdim\tblMeasureWidth>\tblValueWidth\setlength{\tblValueWidth}{\tblMeasureWidth}\fi
\settowidth{\tblMeasureWidth}{Every 2 updates / neither used}\ifdim\tblMeasureWidth>\tblValueWidth\setlength{\tblValueWidth}{\tblMeasureWidth}\fi
\begin{tabular}{>{\raggedright\arraybackslash}p{\dimexpr\linewidth-\tblValueWidth-4\tabcolsep\relax}>{\raggedright\arraybackslash}p{\tblValueWidth}}
\toprule[1pt]
\rowcolor{tblHeader}
\tblhead{Setting} & \tblhead{Value}\\
\midrule[0.4pt]
Starting policy / training questions & Qwen3-4B / 2,460 science questions\\
Optimizer updates / questions per update & 100 / 32\\
Diagnostic / optimization responses & 4 / 8\\
Active guided-sampling allocation & 4 unguided + 4 guided\\
Response cap / generation mode & 3,072 tokens / non-thinking\\
LoRA rank / alpha / dropout & 32 / 32 / 0\\
LoRA targets / bias & All linear layers / none\\
Learning rate / KL coefficient & $10^{-5}$ / $10^{-3}$\\
Trajectory microbatch / gradient clip & 4 / 1.0\\
Training judge & \texttt{gpt-oss-120b}, medium reasoning\\
Judge completion ceiling & 8,192 tokens\\
\midrule[0.4pt]
\rowcolor{tblBaseline}
\multicolumn{2}{l}{\textbf{Additional RuscaRL settings}}\\
Scaffold decay / intra-group allocation & Sigmoid ($\alpha=125$, $t_0=0.2$) / linear\\
Rollout / training contexts & Scaffolded / original question\\
Criterion score levels & 0, 0.5, 1\\
Sampling & Temperature 1; top-$p$ 1; no top-$k$\\
Optimizer / policy clip & AdamW / 0.2\\
Checkpoint interval / capability guidance and routing & Every 2 updates / neither used\\
\bottomrule[1pt]
\end{tabular}
\end{table}

Each RL run processes 3,200 question visits. RSRL generates 25,600 optimization responses and 12,800 diagnostics, excluding initialization. The 256 optimization trajectories per update are processed in microbatches of four before one optimizer step. Runs use one to three A800 GPUs across stages, with judging on separate resources. Equal optimization rollout budgets do not equate the additional computation for the guided proposal's conditional distributions and density evaluation.

\begin{table}[htbp]
\centering
\caption{Fixed diagnosis, exploration, and routing parameters.}
\label{tab:parameters}

\setlength{\tblValueWidth}{0pt}
\settowidth{\tblMeasureWidth}{Value}\ifdim\tblMeasureWidth>\tblValueWidth\setlength{\tblValueWidth}{\tblMeasureWidth}\fi
\settowidth{\tblMeasureWidth}{0.75 / 8}\ifdim\tblMeasureWidth>\tblValueWidth\setlength{\tblValueWidth}{\tblMeasureWidth}\fi
\settowidth{\tblMeasureWidth}{0.9 / 0.8}\ifdim\tblMeasureWidth>\tblValueWidth\setlength{\tblValueWidth}{\tblMeasureWidth}\fi
\settowidth{\tblMeasureWidth}{0.05 / 0.30}\ifdim\tblMeasureWidth>\tblValueWidth\setlength{\tblValueWidth}{\tblMeasureWidth}\fi
\settowidth{\tblMeasureWidth}{1 / 4}\ifdim\tblMeasureWidth>\tblValueWidth\setlength{\tblValueWidth}{\tblMeasureWidth}\fi
\settowidth{\tblMeasureWidth}{0.5 / 0.25}\ifdim\tblMeasureWidth>\tblValueWidth\setlength{\tblValueWidth}{\tblMeasureWidth}\fi
\settowidth{\tblMeasureWidth}{0.5}\ifdim\tblMeasureWidth>\tblValueWidth\setlength{\tblValueWidth}{\tblMeasureWidth}\fi
\settowidth{\tblMeasureWidth}{0.45 / 0.05 / 1.5}\ifdim\tblMeasureWidth>\tblValueWidth\setlength{\tblValueWidth}{\tblMeasureWidth}\fi
\settowidth{\tblMeasureWidth}{5 / 5}\ifdim\tblMeasureWidth>\tblValueWidth\setlength{\tblValueWidth}{\tblMeasureWidth}\fi
\settowidth{\tblMeasureWidth}{$10^{-8}$ / $10^{-6}$}\ifdim\tblMeasureWidth>\tblValueWidth\setlength{\tblValueWidth}{\tblMeasureWidth}\fi
\begin{tabular}{>{\raggedright\arraybackslash}p{\dimexpr\linewidth-\tblValueWidth-4\tabcolsep\relax}>{\raggedright\arraybackslash}p{\tblValueWidth}}
\toprule[1pt]
\rowcolor{tblHeader}
\tblhead{Parameter} & \tblhead{Value}\\
\midrule[0.4pt]
Capability threshold $\tau$ / pooling $\kappa$ & 0.75 / 8\\
Domain retention $\beta_D$ / mapping acceptance & 0.9 / 0.8\\
Guidance activation / saturation deficit & 0.05 / 0.30\\
Guidance $\eta_{\max}$ / contrast clip $b$ & 1 / 4\\
Positive-score threshold / Low fraction & 0.5 / 0.25\\
Partial projection $\xi$ & 0.5\\
Routing $\lambda_{\max}$ / $h_R$ / exponent $p$ & 0.45 / 0.05 / 1.5\\
Criterion / final advantage cap & 5 / 5\\
Normalization $\varepsilon$ / residual tolerance & $10^{-8}$ / $10^{-6}$\\
\bottomrule[1pt]
\end{tabular}
\end{table}

All settings remain fixed during training. With $\kappa=8$, four valid local diagnostics receive one-third of the pooled weight and an available domain prior receives two-thirds.

\Needspace{5\baselineskip}
\subsection{Baseline details}
SFT uses 606 normalized exact-query matches to the training questions, selecting the highest-scoring existing response from a best-of-six source. It trains for one epoch with effective batch size eight, microbatch two, four accumulation steps, learning rate $10^{-5}$, and LoRA rank/alpha 32/32. Loss is restricted to assistant tokens.

Rubric-RL standardizes eligible criteria separately, averages active criterion advantages, and applies root-mean-square scaling and clipping to $[-5,5]$. RSRL instead forms fixed-weight channel advantages before the balanced baseline.

RuscaRL uses linear intra-group scaffolding and sigmoid withdrawal over training progress. Criterion scores are first aggregated into a scalar reward for each response, then standardized across the eight optimization responses. Scaffolds appear during generation; training log probabilities use the original question. Its model, LoRA, judge, learning rate, batch, length cap, and update schedule are given in Table~\ref{tab:training}. It generates four discovery responses per question and uses neither capability-guided exploration nor residual credit routing.

\Needspace{5\baselineskip}
\subsection{Component ablation protocol}
\label{app:ablations}
The comparisons in Section~\ref{sec:ablations} share the frozen capability channels, mappings, and diagnostic procedure. Removing guided exploration sets every optimization group to eight unguided responses, with unit proposal ratios, while retaining residual routing and its guard. Removing residual routing retains guided/unguided sampling and density evaluation but uses $A^0$ for every policy update. Thus the reported contrasts concern the guided proposal and routed credit rule under a fixed capability structure. The shared mapping and adaptive gate are not separately isolated by these variants.

\FloatBarrier
\Needspace{6\baselineskip}
\section{Exploration-strength diagnostics}

\label{app:diagnostics}
\label{sec:diagnostics}
This appendix provides the protocol and complete statistics for the guidance-strength analysis in Section~\ref{sec:analysis}.
Offline diagnostic compares $1\times$, $3\times$, and $4\times$ guidance strength on 1000 triggered questions: 310 with \textsc{Zero} and 690 with \textsc{Low} support. We use the 900 questions valid under all three settings for paired summaries; Scores in Table~\ref{tab:exploration} use the raw 0--1 scale and measure guided-minus-clean target or non-target performance. The effective-signal proxy is a signed sum of guided importance weights, response advantage, and target-score improvement; it is not a gradient norm.

\begin{table}[htbp]
\centering
\caption{Offline guidance diagnostics on 900 common-valid questions. Gains use the 0--1 score scale; ESS is divided by the eight-response group size.}
\label{tab:exploration}

\begin{tabular}{lrrr}
\toprule[1pt]
\rowcolor{tblHeader}
\tblhead{Statistic} & \tblhead{$1\times$} & \tblhead{$3\times$} & \tblhead{$4\times$}\\
\midrule[0.4pt]
Mean target gain & 0.001747 & $-0.004924$ & 0.012816\\
Mean non-target gain & $-0.011115$ & $-0.006001$ & 0.002867\\
Median ESS / 8 & 0.990510 & 0.920261 & 0.865988\\
Mean guided weight share & 0.489837 & 0.427703 & 0.381443\\
Mean effective-signal proxy & 0.044115 & 0.039832 & 0.038713\\
\bottomrule[1pt]
\end{tabular}
\end{table}

Table~\ref{tab:exploration} shows larger target-gain point estimates at $4\times$ together with lower proposal overlap. The paired $4\times-1\times$ gain is 0.011069 with a recorded 95\% bootstrap interval of $[-0.00402,0.02641]$, which includes zero. At $4\times$, fixed-prefix sequence KL and ESS have Spearman correlation $-0.8029$. The separate 50-update comparison has a lower final worst-channel score but higher capability mean at $4\times$ (Table~\ref{tab:early}).

Target and non-target gains compare guided and unguided channel scores. The signed proxy is
\[
U_q=\sum_{i\ \mathrm{guided}}w_iA_i\left(S_{i,c_q^*}-\overline S_{\mathrm{clean},c_q^*}\right).
\]
Here $A_i$ is the advantage used in that diagnostic. Since two negative factors can contribute positively, this statistic is not a beneficial-gradient measure. The paired $4\times-1\times$ target-gain comparison has 450 wins, 250 ties, and 200 losses, with median zero. Fixed-prefix replay gives mean sequence KL values of 0.03658, 0.33031, and 0.58828 at the three strengths.

The 50-update run starts from a fresh base checkpoint and uses the first 50 steps of the training schedule. Table~\ref{tab:early} uses checkpoints. These settings make it an exploratory comparison rather than an isolated strength ablation.

\begin{table}[htbp]
\centering
\caption{Earlier 50-update training comparison. AULCs are normalized by 50. Capability macro averages the six channel scores.}
\label{tab:early}

\begin{tabular}{lrrrrr}
\toprule[1pt]
\rowcolor{tblHeader}
\tblhead{} & \tblhead{Worst} & \tblhead{Final} & \tblhead{Initial weak} & \tblhead{Strong} & \tblhead{Capability}\\
\rowcolor{tblHeader}
\tblhead{Method} & \tblhead{AULC} & \tblhead{worst} & \tblhead{AULC} & \tblhead{$\Delta_{\min}$} & \tblhead{macro}\\
\midrule[0.4pt]
Full $1\times$ & 42.6268 & 42.2695 & 52.6774 & $+0.0620$ & 50.2890\\
Full $4\times$ & 42.3726 & 40.9658 & 52.9860 & $-1.2417$ & 50.8242\\
RuscaRL & 43.2669 & 44.9219 & 53.1383 & $-0.8908$ & 51.1511\\
\bottomrule[1pt]
\end{tabular}
\end{table}

\FloatBarrier
\Needspace{6\baselineskip}
\clearpage
\section{Domain aggregation and routing reliability}
\label{app:record}
Let $\mathcal C_{q,z}^{\mathrm{valid}}$ contain the applicable local channels mapped to $z$ with an available local score and $n_{q,c}>0$. The score and confidence contribution of one question are
\[
O_{q,z}=\frac{\sum_{c\in\mathcal C_{q,z}^{\mathrm{valid}}}r_{q,c}S_{q,c}}{\sum_{c\in\mathcal C_{q,z}^{\mathrm{valid}}}r_{q,c}},\qquad
D_{q,z}=\frac{\sum_{c\in\mathcal C_{q,z}^{\mathrm{valid}}}r_{q,c}m_{q,c}^{\mathrm{conf}}\min(1,r_{q,c})}{\sum_{c\in\mathcal C_{q,z}^{\mathrm{valid}}}r_{q,c}}.
\]
With unit local reliability these become the mean channel score and mean mapping confidence. Mapping confidence does not weight the score average.

Let $\mathcal Q_z^0$ be the initialization questions with an available contribution to $z$. Shared unguided diagnostics initialize
\[
U_{z,0}=\frac1{|\mathcal Q_z^0|}\sum_{q\in\mathcal Q_z^0}O_{q,z},\qquad
d_z^{\mathrm{conf}}=\frac1{|\mathcal Q_z^0|}\sum_{q\in\mathcal Q_z^0}D_{q,z}.
\]
During training, $\bar U_{z,s}$ averages the current $O_{q,z}$ equally over observed questions and updates the domain score through Equation~\eqref{eq:domain_update}. It uses local diagnostic observations rather than prior-pooled scores. Unobserved capabilities retain their previous estimates. Domain and mapping confidence remain fixed.

With both local evidence and a mapped domain estimate, routing uses
\begin{equation}
\chi_{q,c}=\clip\!\left[\nu_{q,c}\min(1,r_{q,c})+(1-\nu_{q,c})d_{m_q(c)}^{\mathrm{conf}}m_{q,c}^{\mathrm{conf}},0,1\right],\qquad
\nu_{q,c}=\frac{n_{q,c}}{n_{q,c}+\kappa}.
\end{equation}
With local evidence alone, $\chi_{q,c}=\clip(r_{q,c},0,1)$; with a mapped domain estimate alone, $\chi_{q,c}=d_{m_q(c)}^{\mathrm{conf}}m_{q,c}^{\mathrm{conf}}$. Channels with neither source are excluded from target selection. The selected channel supplies $\chi_q$. A prior alone does not satisfy the local support test for guided exploration. This composite weight controls routing strength; unit local reliability is a configured weight, not perfect judge accuracy.

\end{document}